\documentclass[authoryear,preprint,12pt]{elsarticle}

\usepackage{soul}
\usepackage{amssymb}
\usepackage{lipsum}
\usepackage[colorlinks=true,citecolor=blue]{hyperref}
\usepackage{lineno}
\usepackage[dvipsnames]{xcolor}
\usepackage{graphicx}
\usepackage{subcaption}
\usepackage{amsmath}
\usepackage{algorithm}
\usepackage{algpseudocode}
\usepackage{array}
\usepackage{tabularx}
\usepackage{amsthm}
\usepackage{subcaption}
\usepackage{tikz}
\usetikzlibrary{arrows.meta, positioning, shapes.geometric}
\usepackage{xltabular}
\newcolumntype{Z}[1]{>{\hsize=#1\hsize}X}
\usepackage{threeparttable} % For table notes
\usepackage{booktabs}
\usepackage{multirow}
\usepackage{ragged2e}
\usepackage{comment}
\usepackage[top=2.5cm, bottom=2.5cm, left=2.5cm, right=2.5cm]{geometry}
\makeatletter
\def\ps@pprintTitle{%
	\let\@oddhead\@empty
	\let\@evenhead\@empty
	\def\@oddfoot{}%
	\let\@evenfoot\@oddfoot}
\makeatother

\begin{document}

\begin{frontmatter}

    \title{A Comprehensive Evaluation Framework for Synthetic Trip Data Generation in Public Transport}
% \title{Toward a Comprehensive Evaluation Framework for Assessing Representativeness, Privacy, and Utility of Synthetic Public Transit Trips}

        \author[rvt]{Yuanyuan Wu}
        \ead{yuanwu@kth.se}

        \author[rvt]{Zhenlin Qin}
        \ead{zhenlinq@kth.se}

 % %        \author[rvt2]{Leizhen Wang}
	% % \ead{leizhen.wang@monash.edu}

 % %        \author[rvt3]{Xiaolei Ma}
 % %        \ead{xiaolei@buaa.edu.cn}

        \author[rvt1]{Zhenliang Ma\corref{cor1}}
        \ead{zhema@kth.se}

    \cortext[cor1]{Corresponding author at: Brinellvägen 23, SE-100 44 Stockholm, Sweden. Tel.:+46 8 790 48 14}
    \address[rvt]{Division of Transport Planning, Department of Civil and Architectural Engineering, KTH Royal Institute of Technology, Stockholm, Sweden}
    \address[rvt1]{Transport Planning Division and Digital Futures, KTH Royal Institute of Technology, Stockholm, Sweden}
 %    \address[rvt2]{Department of Data Science and AI, Monash University, Clayton, Australia}
 %    \address[rvt3]{School of Transportation Science and Engineering, Beihang University, Beijing 100191, China}

    \begin{abstract}
        Synthetic data offers a promising solution to the privacy and accessibility challenges of using smart card data in public transport research. Despite rapid progress in generative modeling, there is limited attention to comprehensive evaluation, leaving unclear how reliable, safe, and useful synthetic data truly are. Existing evaluations remain fragmented, typically limited to population-level representativeness or record-level privacy, without considering group-level variations or task-specific utility.
        To address this gap, we propose a \textit{Representativeness-Privacy-Utility} (\texttt{RPU}) framework that systematically evaluates synthetic trip data across three complementary dimensions and three hierarchical levels (record, group, population). The framework integrates a consistent set of metrics to quantify similarity, disclosure risk, and practical usefulness, enabling transparent and balanced assessment of synthetic data quality.
        We apply the framework to benchmark twelve representative generation methods, spanning conventional statistical models, deep generative networks, and privacy-enhanced variants. Results show that synthetic data do not inherently guarantee privacy and there is no “one-size-fits-all” model, the trade-off between privacy and representativeness/utility is obvious. Conditional Tabular generative adversarial network (CTGAN) provide the most balanced trade-off and is suggested for practical applications. 
        The \texttt{RPU} framework provides a systematic and reproducible basis for researchers and practitioners to compare synthetic data generation techniques and select appropriate methods in public transport applications.
    \end{abstract}

    %Research highlights
    
    \begin{highlights}
        \item A comprehensive \texttt{RPU} (\textit{Representativeness-Privacy-Utility}) framework is proposed to evaluate synthetic trip data in public transport. 
        \item The framework assesses synthetic data across three dimensions and three hierarchical levels (record, group, population).
        \item Twelve representative generation methods are benchmarked, including statistical, deep generative, and privacy-enhanced approaches.
        \item Results show that synthetic data do not inherently guarantee privacy, and no `one-size-fits-all' model exists.
        \item CTGAN achieves the most balanced trade-off between representativeness, privacy, and utility, making it suitable for practical applications.
    \end{highlights}

\end{frontmatter}

%\linenumbers 

\section{Introduction}\label{introduction}
    
    Smart card data have become essential for understanding travel behavior, demand patterns, and policy impacts in public transit systems \citep{pelletier2011smart}. These data provide rich, passively collected records of passenger movements, enabling large-scale and deep empirical analysis. However, growing concerns over data privacy have led public transport agencies and data owners hesitate to share data or apply anonymization techniques such as masking, noise injection, or data swapping. While these methods reduce disclosure risk, they often significantly degrade the utility of the data for the downstream analysis and decision-making \citep{kieu2023synthetic}.

    In response, synthetic data generation using using machine learning and deep generative models (DGMs) has gained growing attention \citep{choi2025gentle}. 
    Techniques such as generative adversarial networks (GANs) \citep{goodfellow2014generative} and variational autoencoders (VAEs) \citep{huang2019variational} strive to reproduce the statistical properties of real data, while also balancing the protection of sensitive or identifiable information. 

    Despite growing interest in synthetic data, most research has focused on developing generative models, with relatively limited attention to evaluating whether the generated data are both useful and privacy-preserving. A systematic and robust evaluation framework remains remains underdeveloped. Such a framework is critical for both researchers and practitioners to assess whether synthetic data are fit for purpose in real-world transport applications. In its absence, model validation, comparison, and practical adoption remain ad hoc and inconsistent.
    Evaluating synthetic data requires answering three fundamental questions: 
    \begin{enumerate}
        \item Does the data accurately reflect the real-world distributions?
        \item Does it sufficiently protect sensitive or identifiable information?
        \item Is it still useful for downstream analysis and decision-making?
    \end{enumerate}
    
    These questions correspond to three core dimensions of synthetic data quality that we refer to as: \textit{representativeness}, \textit{privacy}, and \textit{utility}.
    However, current evaluation practices in the literature remain limited in several important respects. First, assessments are heavily skewed toward representativeness, typically measured through distributional similarity at the population level (e.g., marginal or joint distributions) \citep{kapp2023generative}. Individual and group level representativeness, which are critical for capturing heterogeneous travel behaviors and subgroup dynamics, are often overlooked. Second, privacy evaluations are narrowly focused, usually limited to duplicate checks or record-level membership inference attacks (MIA) \citep{smolak2020population}. Few studies consider group-level privacy, such as leakage of sensitive signals from identifiable subpopulations, or population-level privacy, which concerns the inference of aggregate patterns or behavioral trends. Third, utility is often conflated with representativeness and assessed using similarity metrics. While some degree of similarity is necessary for utility, relying solely on distributional similarity provides an incomplete picture of whether the synthetic data are fit for downstream use. The true utility of synthetic data requires dedicated evaluation, particularly for tasks such as monitoring, planning, or policy analysis. 
    
    Moreover, the conflation of metrics across evaluation dimensions is problematic. For example, high distributional similarity is sometimes interpreted as both good representativeness/utility and low privacy risk \citep{berke2022generating}. It is an assumption that does not always hold. Such mixed or overlapping use of metrics can lead to misleading conclusions and hinders rigorous comparison between models or informed selection for deployment.

    This paper addresses these limitations by proposing a multi-dimensional, multi-level evaluation framework for synthetic public transport data, based on dedicated indicators (Figure~\ref{fig:diagram}). 
    % Specifically, we assess 
    % \begin{itemize}
    %     \item \textbf{Representativeness} at record-level plausibility, group-level diversity, and population-level fidelity;
    %     \item \textbf{Privacy} in terms of record-level disclosure risk, group-level exposure, and population-level leakage; and
    %     \item \textbf{Utility} by evaluating whether synthetic data can effectively replace real data in downstream tasks relevant to monitoring, planning, and policy analysis.
    % \end{itemize}
    We apply this framework to benchmark a diverse set of statistical, deep generative, and privacy-enhanced models, using real-world AFC trip data. Empirical insights into generative model performance and the trade-offs involved in producing synthetic trip data for public transport applications are derived. 

    The remainder of the paper is organized as follows. Section \ref{sec:relatedwork} reviews related work. 
    Section \ref{sec:framework} develops the evaluation framework. Section~\ref{sec:models} describes the generative models for benchmarking. Section \ref{sec:case} demonstrates its use for synthetic trip data generation in public transport. Section \ref{sec:conclusion} summarizes the key findings and future research.

\section{Related work} \label{sec:relatedwork}
  \subsection{Background}
    % AFC systems have become a ubiquitous component of modern public transit networks, recording vast volumes of passenger transactions through smart cards, mobile devices, or contactless payment methods. These systems, initially implemented primarily for fare collection efficiency, have evolved into rich data sources that provide unprecedented insights into transit operations and passenger behavior \citep{pelletier2011smart, ma2013mining}, such as origin-destination estimation \citep{bwambale2019modelling}, peak-hour demand identification and management \citep{halvorsen2020demand}, passenger response to intervention \citep{mo2022inferring}, and evaluation of fare policies \citep{wu2025data}. 
    
    % Despite the analytical value of AFC data, its use is constrained by privacy regulations, institutional data sharing policies, and concerns over the potential re-identification of individuals \citep{ioannou2020privacy}. Privacy remains a critical issue in publishing mobility data. It is shown that even aggregated mobility data can expose sensitive trajectory information and lead to privacy breaches \citep{xu2017trajectory}, while deep learning models can be vulnerable to membership inference attacks \citep{shokri2017membership}. These challenges have motivated interest in privacy-preserving alternatives, the use of synthetic data has become increasingly common in transportation research, particularly in addressing privacy concerns associated with large-scale mobility datasets \citep{da2025generative}. 
    
    Synthetic data refers to artificially created records that mimic patterns and structures observed in real-world datasets \citep{drechsler2011synthetic}. These datasets are typically produced by fitting generative models to the original data in order to capture its key statistical characteristics, either the joint probability $p(x,y)$ or the marginal distribution $p(x)$. Once trained, such models can be used to generate new data instances, such as travel trajectories, that resemble the original distribution while allowing flexible scaling \citep{goncalves2020generation}. A wide range of generative modeling techniques have been explored in the literature, ranging from classical probabilistic methods to modern deep learning-based approaches, representing as Deep Generative Models (DGMs) \citep{choi2025gentle}. 
    % The primary distinction between these two categories lies in their reliance on predefined assumptions about the data distribution, such as Gaussianity or conditional independence. 
    
    Classical models, such as Gaussian mixture models \citep{reynolds2015gaussian}, Gaussian copulas \citep{nelsen2006introduction}, and Bayesian networks (BNs) \citep{koller2009probabilistic}, fit explicit probabilistic distributions directly to real data.
    While these models are interpretable and effective in many applications, they typically rely on simplifying assumptions (e.g., Gaussianity, conditional independence) and may struggle to capture complex, high-dimensional, or non-linear patterns in real-world data such as mobility behaviors.
    Nevertheless, they provide strong and transparent baselines for evaluating the performance of DGMs. 
    Moreover, their underlying principles continue to influence modern architectures. For example, GMMs are used in Conditional Tabular GANs (CTGAN) to sample training batches and mitigate mode collapse \citep{xu2019modeling}; copula-based normalization has been proposed to improve model transferability across domains \citep{jutras2024copula}; and BNs have been leveraged to derive topological orderings that guide the autoregressive generation process in large language models (LLMs) for population synthesis \citep{lim2025large}.

    %DGMS related to this study review. 
    Unlike classical models, DGMs leverage neural network architectures to flexibly approximate non-linear dependencies and latent structure in the data without relying on strong parametric assumptions. Two widely used families of DGMs for generating mobility data are Generative Adversarial Networks (GANs) \citep{goodfellow2014generative} and Variational Autoencoders (VAEs) \citep{kingma2013auto}.
    GANs consist of a generator and a discriminator trained in opposition: the generator aims to produce realistic synthetic data, while the discriminator attempts to distinguish between real and synthetic samples. Variants such as CTGAN adapt this framework for tabular data by conditioning on specific feature values and using training tricks to reduce mode collapse \citep{xu2019modeling}. Extensions like PATE-GAN incorporate differential privacy by training the discriminator on an ensemble of teacher models with noise injection to protect individual data contributions \citep{jordon2018pate}. 
    VAEs are probabilistic models that encode data into a latent space and reconstruct samples by decoding from this space \citep{kingma2013auto}. They optimize a variational lower bound on the data likelihood, balancing reconstruction accuracy with latent space regularization. Compared to GANs, VAEs are generally more stable to train and offer a structured, interpretable latent space \citep{doersch2016tutorial}. However, they may produce less sharp or over-regularized samples if not carefully tuned \citep{lucic2018gans}, especially for categorical or structured data \citep{bao2022variational}.

    In recent work, more expressive architectures have been proposed. For instance, diffusion models iteratively denoise random noise toward realistic samples through learned reverse processes, achieving state-of-the-art performance in image synthesis and increasingly adapted for tabular data generation \citep{croitoru2023diffusion,yang2023diffusion}. Normalizing flows transform a simple distribution into a complex one through a sequence of invertible mappings, allowing exact likelihood estimation \citep{papamakarios2021normalizing}. Large Language Models (LLMs) have also been explored as autoregressive generators for structured records, leveraging token-level conditioning and fine-tuning strategies \citep{patel2024datadreamer,lim2025large}.
    
    %These models offer diverse strengths but also present challenges, such as mode collapse in GANs, limited interpretability in VAEs, or scalability concerns in diffusion models \citep{zhang2024chattraffic}, which motivate systematic, multi-dimensional evaluation frameworks like the one proposed in this study.

    % refer to broad literature review. 
    DGMs have been widely applied across various urban mobility data generation tasks, including population synthesis \citep{badu2022composite,kim2023deep}, synthetic traffic data \citep{wang2020seqst,li2022spatial}, origin-destination (OD) matrices generation \citep{rong2023complexity,rong2023origin}, and vehicle or human trajectory modeling \citep{huang2019variational,fontana2023gans}, among others. For broader overviews, readers are referred to the reviews by \cite{kapp2023generative} and \cite{choi2025gentle}.  
    % In this work, we focus specifically on the use of DGMs for synthesizing smart card (AFC) data, as detailed in the following review.

  \subsection{DGMs for smart card data synthesis}
    In the context of smart card data synthesis, GANs are the most commonly used models for smart card data synthesis, while VAEs often serve as baseline benchmarks \citep{badu2022composite}. For example, \cite{kim2022imputing} used a conditional GAN to impute qualitative attributes of trip chains from smart card data, showing improved fidelity, diversity, and creativity compared to a conditional VAE and other baselines, as measured by Jensen-Shannon Distance (JSD) and confusion matrices. Privacy concern was not included in their evaluation. 
    
    Focusing more directly on synthetic trip generation, \cite{kieu2023synthetic} utilized a tabular GAN (CTGAN) to synthesize bus smart card trip records from South-East Queensland’s transit system. Their study compared the performance of CTGAN with that of a Bayesian network using over one million real trip records. Representativeness was evaluated by examining how well the synthetic data replicated key statistical distributions. While both models captured general travel patterns, CTGAN tended to overestimate peak-hour travel and rare trip types, whereas the Bayesian network provided a closer fit for attributes such as travel time distributions. For privacy evaluation, the authors conducted a duplicate detection check by identifying any synthetic records that exactly matched real trips. Although this method provides a basic indication of overfitting or memorization, it falls short of offering a comprehensive or formal privacy assessment.
    
    To address privacy more rigorously, \cite{badu2020differentially} developed a differentially private (DP) GAN for generating daily travel activity diaries. Their model incorporated differential privacy via DP-SGD during training, thereby ensuring that individual-level information in the training data was protected. In addition to evaluating the representativeness through marginal distributions and pairwise correlations, they employed membership inference attacks \citep{shokri2017membership} to quantify the level of privacy risk. While their study focused on household travel surveys (HTS) data, the approach is adaptable to smart card data synthesis.
    
    HTS datasets provide rich behavioral and socioeconomic information, making them valuable for modeling urban mobility. However, their utility is constrained by several well-known limitations: they are expensive and time-consuming to collect, suffer from recall bias, are infrequently updated, and are increasingly restricted by privacy concerns. 
    To address these challenges, synthesizing smart card (SC) data using generative models offers a promising alternative, as SC data provide large-scale, high-resolution observations collected passively over time \citep{uugurel2024learning}. Recent studies have explored data fusion strategies that combine the complementary strengths of SC data. For example, \cite{xu2024generative} proposed an GAN-based framework to infer the age attribute extracted from SC data and examine the relationship between built environment and age using features from Points of Interest data. 
    \cite{vo2025novel} proposed a cluster-based data fusion method that leverages SC data to enhance the spatial granularity of HTS datasets, which are limited by low sampling rates. Similarly, \citet{lee2025collaborative} introduced a collaborative GAN-based framework to fuse HTS and SC data for generating detailed activity schedules. While these works evaluate the representativeness and utility of the synthesized data, they do not include formal privacy assessments.

  \subsection{Synthetic data evaluation}
   
    %uniqueness testing: estimating how unique that individual’s mobility data are with respect to the mobility data of the other individuals represented in the dataset. 
    
    Privacy is commonly stated in the literature as the main motivation for applying DGMs to mobility data synthesis. However, in practice, many studies focus primarily on demonstrating the feasibility or novelty of their generative approaches, with evaluation metrics tailored to specific modeling goals. To demonstrate the effectiveness of proposed DGMs in reproducing diverse population, \cite{kim2023deep} used standardized root mean square error (SRMSE) for marginal and bivariate distributions, and confusion metrics (precision, recall, F1) to indicate feasibility and diversity. \cite{zhang2023csgan} assessed modality-aware trajectory generation via Jensen-Shannon Divergence (JSD) on modality distributions and transitions, while \cite{kieu2023synthetic} used the Wasserstein distance to compare spatial distributions.
    In benchmarking CTGAN, TVAE, and Gaussian Copulas for carsharing trip synthesis, \cite{albrecht2024fake} evaluated fidelity through statistical tests (Kolmogorov-Smirnov, Chi-squared) and graphical comparisons of marginal distributions and pairwise correlations. These metrics help quantify how well the synthetic data preserve the structural characteristics of the real data, especially for tabular data. 
   % Privacy evaluation is frequently overlooked in these studies. To illustrate this imbalance across the literature, 
   
    Privacy concerns are frequently addressed in studies on trajectory generation. For example, \cite{kulkarni2018generative} benchmarked RNNs, GANs, and Gaussian copulas in generating synthetic mobility trajectories, evaluating the privacy–utility tradeoff using membership inference attacks (MIA). Similarly, \cite{fontana2023gans} introduced a privacy-aware framework that combines a GAN for trajectory generation with a deep learning-based user identification model. Privacy was evaluated through identification accuracy, following the data uniqueness concept proposed by \cite{sweeney2002k}. More recently, \cite{zhang2025noise} proposed a diffusion model incorporating collaborative noise priors for urban mobility synthesis. They evaluated the model using marginal distribution alignment, privacy protection via uniqueness testing and MIA, and utility through downstream prediction performance.   

    Overall, most existing studies assess the quality of synthetic data using only one or two dimensions, typically focusing on either \textit{statistical similarity} (e.g., distributions, pairwise distances) or \textit{utility} in downstream tasks (e.g., prediction accuracy). Common fidelity measures such as Jensen-Shannon Divergence (JSD) and Wasserstein distance are widely applied to evaluate how well the synthetic data mirrors real data at distribution/population-level. Frameworks like SDV \citep{patki2016synthetic} support benchmarking across statistical models and GAN-based generators, with an emphasis on utility and general similarity metrics. In contrast, privacy is often overlooked, and few studies comprehensively evaluate synthetic data across representativeness, privacy, and utility, especially at multiple levels.
    Recent efforts have begun to adopt more multi-dimensional evaluation strategies. For example, \cite{platzer2021holdout} introduced a holdout-based fidelity and privacy assessment for mixed-type tabular data using individual distance-to-nearest-record as a proxy for privacy risk. \cite{eigenschink2023deep} identified five key criteria for evaluating synthetic data quality: representativeness, novelty, realism, diversity, and coherence. Similarly, \cite{lautrup2025syntheval} proposed SynthEval, an open-source framework integrating fidelity and privacy metrics across both categorical and numerical features for unified benchmarking. 
    However, despite these advances, evaluations are often limited to the population/distribution level, neglecting how synthetic data performs at finer granularities such as individual records or within specific user groups. 
    %This limits the ability to detect nuanced biases, privacy risks, or functional inconsistencies—especially in applications where user-level behavior and subgroup patterns matter, such as in smart card-based transit data. Few studies offer a structured evaluation across both dimensions and levels, leaving a gap in holistic assessment practices.
    
    Table~\ref{tab:related_work_dgms} provides an overview of representative studies on \textit{synthetic mobility data} evaluation, checking their focus on representativeness, utility, and privacy. As shown, most works prioritize representativeness at population-level and downstream utility, while systematic privacy assessment remains limited. This imbalance highlights the need for a more comprehensive and structured evaluation across both dimensions and levels.

    % refine the table at different levels.
    
    \begin{table}[ht]
        \centering
        \scriptsize
        \caption{Summary of related work applying DGMs for synthetic mobility data}
        \resizebox{\textwidth}{!}{%
        \renewcommand{\arraystretch}{1.2}
        \begin{tabular}{|c|c|c|c|l|c|c|c|}
            \hline
            \multirow{2}{*}{\textbf{Study}} & \multirow{2}{*}{\textbf{Model Type}} & \multirow{2}{*}{\textbf{Benchmark(s)}} & \multirow{2}{*}{\textbf{Data Type}} & \multirow{2}{*}{\textbf{Use Case} }& \multicolumn{3}{c|}{\textbf{Evaluations}} \\\cline{6-8}
              & & & & &\textbf{R$^*$}&\textbf{P$^*$} &\textbf{U$^*$}\\
            \hline
            \cite{kulkarni2018generative} & GAN & Copulas, RNNs & GPS & generate trajectory&  $\checkmark$ & $\checkmark$ & $\times$\\
            \hline
            \cite{borysov2019generate} & VAE & Gibbs sampler, BNs & HTS & synthesize population& $\checkmark$ & $\times$  & $\times$\\
            \hline
            \cite{badu2020differentially} & DP-GAN & & HTS $+$ GPS & synthesize travel diary & $\checkmark$ & $\checkmark$&$\times$\\
            \hline
            \cite{kim2022imputing} & cGAN &Gibbs sampler, BN, VAE & SC $+$ HTS & impute sociodemographic attributes and trip purposes & $\checkmark$ & $\times$ & $\times$ \\
            \hline
            \cite{kieu2023synthetic} & CTGAN & BN & SC & synthesize trips &$\checkmark$ & $\checkmark$ & $\times$\\
            \hline
            \cite{kim2023deep} & WGAN & BN, VAE & HTS & synthesize population & $\checkmark$ & $\times$ & $\times$\\
            \hline
            \cite{zhang2023csgan} & GAN & other GANs & GPS & generate trajectory & $\checkmark$ & $\times$ & $\times$\\
            \hline
            \cite{fontana2023gans} & privacy-aware GAN & N.A. & GPS & generate trajectory & $\checkmark$ & $\checkmark$ & $\checkmark$ \\
            \hline
            \cite{albrecht2024fake} & Multiple & CTGAN, TVAE, Copula & carsharing data & synthetic tabular data & $\checkmark$ & $\times$ & $\times$\\
            \hline
            \cite{xu2024generative} & GAN & Random Forest, BN, VAE & SC $+$ PoI & infer socioeconomic characteristics of passengers & $\checkmark$ & $\times$ & $\times$\\
            \hline
            \cite{lee2025collaborative} & GAN & & SC $+$ HTS & generate activity schedules & $\checkmark$ & $\times$ &$\checkmark$ \\
            \hline
            \cite{zhang2025noise} & Diffusion & GANs, Diffusion & GPS & generate trajectory & $\checkmark$ & $\checkmark$ & $\checkmark$\\
            \hline
            \textbf{This study} & Multiple &GANs, VAEs, etc. & SC & synthetic trips & $\checkmark$ & $\checkmark$ & $\checkmark$ \\
            \hline
            \multicolumn{8}{l}{$^*$R: Representativeness; P: Privacy; U: Utility}
        \end{tabular}
        }
        \label{tab:related_work_dgms}
    \end{table}

\section{\texttt{RPU} evaluation framework}\label{sec:framework}
\begin{figure}[ht]
         \centering
         \includegraphics[width=\linewidth]{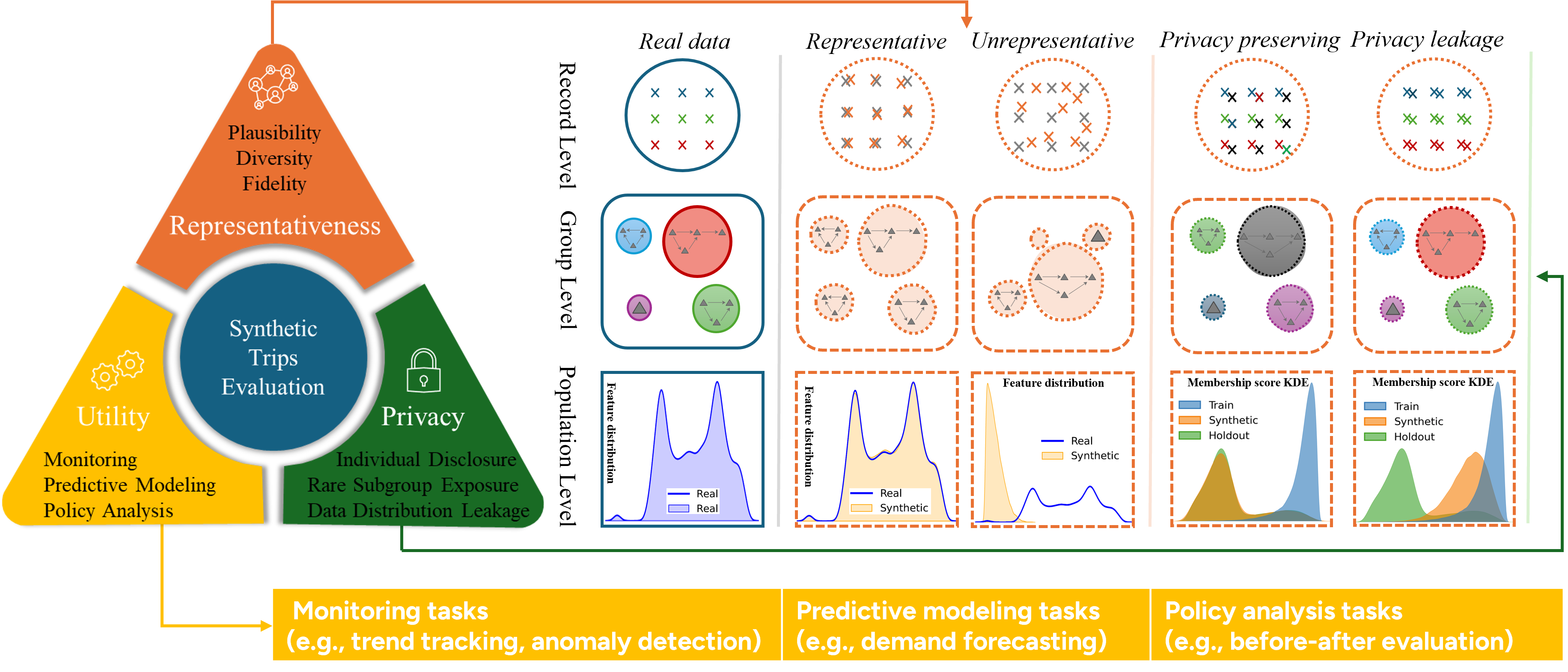}
         \caption{Representativeness, privacy, utility evaluation framework for synthetic trips.}
         \label{fig:diagram}
    \end{figure}

    To comprehensively assess the quality of synthetic data, we propose a structured evaluation framework organized along three key dimensions: \emph{representativeness}, \emph{privacy}, and \emph{utility}, denoted as \texttt{RPU} framework (Figure~\ref{fig:diagram}, Table~\ref{tab:eval_RPU}).
    Each dimension is assessed through multiple aspects, reflecting distinct perspectives of data quality.
            
    \begin{table}[htbp]
    \centering
    \caption{Definition and indicators of \texttt{RPU} evaluation framework}
    \scriptsize
    \renewcommand{\arraystretch}{1.4}
    \setlength{\tabcolsep}{5pt}
    \begin{tabularx}{\textwidth}{>{\raggedright\arraybackslash}p{2cm} 
                                 >{\raggedright\arraybackslash}p{10cm} 
                                 >{\raggedright\arraybackslash}p{4cm}}
        \toprule
        \multicolumn{3}{l}{\textbf{Representativeness}}\\
        \textbf{Aspects} & \textbf{Definition} & \textbf{Indicators} \\
        \midrule
        \multirow{3}{2cm}{Record-level plausibility}
        & \multirow{3}{10cm}{The extent to which each synthetic data record exhibits plausible, coherent, and realistic behavior when considered as a stand-alone entity.}
        & - Visual inspection\\
         & &- Rule-based logic validation\\
         & &- Expert judgment \\
    
         \multirow{3}{2cm}{Group-level heterogeneity}
        & \multirow{3}{10cm}{The degree to which the synthetic data preserves the distinctive behavioral patterns and statistical properties of subpopulations defined by shared characteristics.}
        & \multirow{3}{4cm}{- KLD, JSD, EMD across groups}\\
        & & \\
        & & \\
         \multirow{3}{2cm}{Population-level similarity}
        & \multirow{3}{10cm}{The extent to which the overall statistical structure of the synthetic data matches that of the original dataset, including marginal distributions, joint feature dependencies, and spatial-temporal patterns.}
        &  \\
        & &- KLD, JSD, EMD \\
        & & \\\midrule
        \multicolumn{3}{l}{\textbf{Privacy}}\\
        \textbf{Aspects} & \textbf{Definition} & \textbf{Indicators} \\
        \midrule
        \multirow{3}{2cm}{Record-level disclosure}
        & \multirow{3}{10cm}{The risk that an individual synthetic record reveals the membership or attributes of a particular real individual.}
        & - Visual inspection\\
         & &- MIA\\
         & & - Attribute inference \\
    
         \multirow{3}{2cm}{Group-level exposure}
        & \multirow{3}{10cm}{The risk of reproducing rare or vulnerable groups, revealing group-level patterns that could be exploited, or amplifying the exposure of marginalized populations.}
        & \multirow{3}{4cm}{- K-NN distance ratio across groups}\\
        & & \\
        & & \\
         \multirow{3}{2cm}{Population-level leakage}
        & \multirow{3}{10cm}{The risk that synthetic data may leak sensitive structural,  behavioral, and societal patterns of the underlying population.}
        &  \\
        & &- K-NN distance ratio \\
        & & \\ 
        \midrule
        \multicolumn{3}{l}{\textbf{Utility}}\\
        \textbf{Aspects} & \textbf{Definition} & \textbf{Indicators} \\
        \midrule
        \multirow{3}{2cm}{Monitoring}
        & \multirow{3}{10cm}{Use synthetic data to track trends, detect anomalies, and support operational dashboards.}
        & - Time-series similarity\\
         & &- Peak location accuracy \\
         & & - Change-point detection  \\
    
         \multirow{3}{2cm}{Predictive modeling}
        & \multirow{3}{10cm}{Use synthetic data to train or evaluate models that predict future or unknown outcomes.}
        & - TSTR  \\
        & & - Accuracy, AUC, F1\\
        & & - Feature importance correlation\\
         \multirow{3}{2cm}{Policy analysis}
        & \multirow{3}{10cm}{Use synthetic data to simulate or evaluate effects of interventions or interruptions.}
        &  \\
        & & - Expert evaluation  \\
        & & \\
        \bottomrule
        \end{tabularx}
        \label{tab:eval_RPU}
    \end{table}
    
    % Each dimension is further decomposed into distinct aspects evaluated at the record, group, and population levels. The representativeness dimension captures the extent to which synthetic data replicates the statistical characteristics of the original dataset, including record-level plausibility, subgroup heterogeneity, and population-level similarity. The privacy dimension addresses the potential risks of disclosing sensitive information, operationalized through individual-level disclosure risks, group-level exposure, and population-level structural leakage. The utility dimension reflects the practical value of synthetic data in supporting downstream applications, including system monitoring, predictive modeling, and policy analysis. These aspects are evaluated using a suite of indicators—such as divergence measures (e.g., Kullback–Leibler, Jensen–Shannon, and Wasserstein distances), $k$-anonymity, membership inference attacks, time-series similarity, and transfer learning metrics (e.g., Train on Synthetic, Test on Real). This multi-level, multi-dimensional framework provides a holistic basis for evaluating the fidelity, privacy-preserving properties, and functional applicability of synthetic datasets.
     
% , as shown in Figure~\ref{fig:diagram}.

  \subsection{Representativeness and indicators}\label{subsec:rindicator}
    The representativeness dimension, denoted \( \mathcal{R} \), measures how well the synthetic data replicates key characteristics of the real data. This includes \textit{population-level} similarity in global distributions and structural relationships (\( \mathcal{R}_p \)), \textit{group-level} fidelity in capturing subgroup-specific patterns (\( \mathcal{R}_g \)), and \textit{record-level} plausibility and internal coherence (\( \mathcal{R}_r \)). 

    A fundamental requirement for synthetic data is that its distribution \( P(x') \) should closely approximate the distribution of the original data \( D(x) \). This similarity is typically quantified using statistical divergence or distance measures. One widely used metric is the \textit{Kullback-Leibler divergence (KLD)}, defined as:
    \begin{equation}\label{equ:kld}
    \mathrm{KL}(D(x) \,\|\, P(x')) =
        \begin{cases}
            \sum\limits_{x} D(x) \log \frac{D(x)}{P(x')} & \text{(discrete case)} \\[10pt]
            \int D(x) \log \frac{D(x)}{P(x')} \, dx & \text{(continuous case)}
        \end{cases}
    \end{equation}

    KL divergence measures the information loss when \( P(x') \) is used to approximate \( D(x) \). It is non-negative and equals zero if and only if the two distributions are identical almost everywhere. A \textbf{smaller} KL value indicates higher similarity between the two distributions. KLD is asymmetry and not bounded. The \textit{Jensen-Shannon divergence (JSD)} is often used as a symmetric and smoothed alternative:
    \begin{equation}\label{equ:jsd}
        \begin{split}
            \mathrm{JSD}(D(x) \,\|\, P(x')) &= \frac{1}{2} \mathrm{KL}(D(x) \,\|\, M(x)) 
            + \frac{1}{2} \mathrm{KL}(P(x') \,\|\, M(x)) \\
            &\text{where } M(x) = \frac{1}{2}(D(x) + P(x'))
        \end{split}
    \end{equation}

    JSD is a symmetric and bounded divergence measure. When using log base 2, which is common in information-theoretic contexts, its values lie within the range \([0, 1]\), facilitating interpretability and comparability across models.
    Another widely adopted metric, especially in the evaluation of generative models, is the \textit{Wasserstein distance}, which computes the cost of moving one distribution to match another, with smaller distances indicating closer distributions, is commonly used to evaluate the spatial distribution similarity:
    \begin{equation}
         W(D, P) = \inf_{\gamma \in \Pi(D, P)} \mathbb{E}_{(x, x') \sim \gamma} \left[ \| x - x' \| \right]
    \end{equation}
    where \( \Pi(D, P) \) denotes the set of all joint distributions (couplings) with marginals \( D(x) \) and \( P(x') \). Wasserstein distance, is also known as the \textit{Earth Mover's Distance (EMD)}, provides meaningful gradients even when the distributions do not overlap, making it particularly suitable for training and evaluating DGMs.

    While many existing studies focus primarily on population-level evaluation, assessing \textit{how well the overall statistical structure of synthetic data aligns with the original data} using metrics such as KLD, JSD, and EMD. Such aggregate measures may overlook important variations at finer granularities. Specifically, high distributional similarity does not guarantee that synthetic records are individually plausible or that subpopulation characteristics are faithfully reproduced \citep{eigenschink2023deep}. 

    To provide a more comprehensive assessment, our framework incorporates two additional levels of evaluation. The \textit{group-level} component assesses whether the synthetic data preserves distinct behavioral patterns across meaningful subsets of the data, defined by features such as peak vs. off-peak hours, weekdays vs. weekends, or user-defined criteria. This is measured using conditional KLD, JSD, and EMD. The \textit{record-level} is defined as the degree to which synthetic data retain natural plausibility, even if statistical properties are well-matched. Assessment at this level is performed through visual inspection, rule-based validation, and expert judgment, focusing on the behavioral plausibility of each synthetic trip and the internal consistency of its attributes, such as valid transfer sequences and temporally coherent travel timelines. 

  \subsection{Privacy and indicators}
    We evaluate privacy also at three levels: the \textit{record level} (\( \mathcal{P}_r \)), such as identity and attribute disclosure; the \textit{group level} (\( \mathcal{P}_g \)), including rare subgroup exposure; and the \textit{population level} (\( \mathcal{P}_p \)), including structural or behavioral pattern leakage. 

    To assess record-level privacy risk, we adopt a customized membership inference attack (MIA) tailored to our synthetic data evaluation setting (Algorithm~\ref{alg:mia_rf}), which estimate the probability $p_{mia}(x)$ that a given record $x \in D_{real}$ was part of the training data. Specifically, we train a binary classifier to distinguish between real training records (labeled as members) and holdout records (non-members) based on selected feature vectors. The attack model is then applied to synthetic data, where all records are presumed non-members. We report the mean predicted membership probability across synthetic records, which reflects the average confidence of the classifier that synthetic samples resemble training data. This approach provides a practical and interpretable signal of overfitting or memorization by the generative model. Additionally, we visualize the distributions of membership scores for training, holdout, and synthetic samples using kernel density estimation (KDE) plots, enabling a qualitative comparison of risk exposure across datasets. 
    \begin{algorithm}[H]
        \caption{Membership Inference Attack via Random Forest Classifier}
        \label{alg:mia_rf}
        \begin{algorithmic}[1]
            \Require Real training data $D_{\text{train}}$, real holdout data $D_{\text{holdout}}$, synthetic data $D_{\text{syn}}$, feature list $\mathcal{F}$

            \Function{PrepareMIAData}{$D_{\text{train}}, D_{\text{holdout}}, \mathcal{F}$}
                \State Assign label 1 to $D_{\text{train}}$, label 0 to $D_{\text{holdout}}$
                \State Combine both datasets into $D_{\text{combined}}$
                \State $X \gets D_{\text{combined}}[\mathcal{F}]$, $y \gets D_{\text{combined}}[\text{label}]$
                \State \Return $X, y$
            \EndFunction

            \Function{TrainMIAClassifier}{$X, y$}
                \State Split $X, y$ into training and test sets with stratified sampling
                \State Train Random Forest classifier $\mathcal{C}$ on training set
                \State Predict membership scores $y_{\text{pred}}$ on test set
                \State Compute AUC: $\text{AUC} \gets \text{roc\_auc}(y_{\text{test}}, y_{\text{pred}})$
                \State \Return $\mathcal{C}, \text{AUC}$
            \EndFunction

            \Function{EvaluateOnSynthetic}{$\mathcal{C}, D_{\text{syn}}, \mathcal{F}$}
                \State $X_{\text{syn}} \gets D_{\text{syn}}[\mathcal{F}]$
                \State Compute membership scores: $s \gets \mathcal{C}.predict\_proba(X_{\text{syn}})[:, 1]$
                \State \Return Mean confidence: $\mu_s \gets \text{mean}(s)$
            \EndFunction

            \Function{PlotMIAScoreDistribution}{$\mathcal{C}, D_{\text{train}}, D_{\text{holdout}}, D_{\text{syn}}, \mathcal{F}$}
                \State Compute scores $s_{\text{train}}, s_{\text{holdout}}, s_{\text{syn}}$ using $\mathcal{C}$ on each dataset
                \State Combine scores with labels into one data structure
                \State Plot KDE of score distributions for each type
            \EndFunction
        \end{algorithmic}
    \end{algorithm}

    Divergence-based measures (e.g. KLD, JSD, and EMD) are commonly used in the literature to assess the population or sub-group level leakage, since overly close distribution matches imply privacy risks. However, because these measures are already employed to evaluate representativeness in our framework, we instead evaluate population- and group-level privacy risks using a \textit{k-nearest neighbor} (k-NN) distance-based analysis (Algorithm~\ref{alg:knn_privacy}). At the population level, we compute the average distance from each synthetic record to its (\( k\)) nearest real records, and compare this to the average distance between each real record and its (\( k+1\)) nearest real neighbors (excluding the self-match). The ratio between these two quantities serves as an interpretable metric of memorization or overfitting risk. A lower ratio indicates that synthetic records are more tightly clustered around real data than expected, suggesting potential privacy leakage. 

    To extend this analysis to the group level, we stratify the data by a grouping variable (\(\mathcal{F}\)) and compute the k-NN distance ratio within each group. This group-wise analysis captures variations in privacy risk across different segments of the data distribution. %The mean of group-wise k-NN distance ratios is used as the summary statistic to evaluate group-level privacy risk. 
    \begin{algorithm}[H]
        \caption{k-NN Distance-Based Privacy Evaluation}
        \label{alg:knn_privacy}
        \begin{algorithmic}[1]
            \Require Real data $D_\text{real}$, Synthetic data $D_\text{syn}$, Feature list $\mathcal{F}$, Optional group column $g$, Number of neighbors $k$, Thresholds $(\theta_\text{low}, \theta_\text{high})$

            \Function{EvaluatePopulationLevelPrivacy}{$D_\text{real}, D_\text{syn}, \mathcal{F}, k$}
                \State Fit k-NN model on $D_\text{real}[\mathcal{F}]$
                \State Compute distances from each synthetic record to $k$ nearest real records: $d_{\text{syn} \rightarrow \text{real}}$
                \State Compute mean over all distances: $\bar{d}_{\text{syn} \rightarrow \text{real}}$
    
                \State Fit k+1-NN model on $D_\text{real}[\mathcal{F}]$
                \State Compute distances from each real record to its $k+1$ nearest real neighbors
                \State Exclude self-distance and compute mean: $\bar{d}_{\text{real} \rightarrow \text{real}}$
    
                \State Compute distance ratio: $\rho = \frac{\bar{d}_{\text{syn} \rightarrow \text{real}}}{\bar{d}_{\text{real} \rightarrow \text{real}}}$
                \State \Return $\rho$
            \EndFunction

            \Function{EvaluateGroupLevelPrivacy}{$D_\text{real}, D_\text{syn}, \mathcal{F}, g, k$}
                \State Extract unique group values: $G \gets \text{unique}(D_\text{real}[g])$
                \State Initialize list $\mathcal{R} \gets []$  \Comment{To store group-wise ratios}
                \ForAll{$v \in G$}
                    \State $D_\text{real}^{(v)} \gets$ real records with $g = v$
                    \State $D_\text{synth}^{(v)} \gets$ synthetic records with $g = v$
                    \If{$|D_\text{real}^{(v)}| > k$ \textbf{and} $|D_\text{syn}^{(v)}| > 0$}
                        \State Compute $\rho^{(v)} \gets$ \Call{EvaluatePopulationLevelPrivacy}{$D_\text{real}^{(v)}, D_\text{syn}^{(v)}, \mathcal{F}, k$}
                        \State Append $\rho^{(v)}$ to $\mathcal{R}$
                    \Else
                        \State Skip group $v$ due to insufficient data
                    \EndIf
            \EndFor
            \State \Return Group-level mean ratio: $\bar{\rho}_\text{group} = \text{mean}(\mathcal{R})$
            \EndFunction
        \end{algorithmic}
    \end{algorithm}

  \subsection{Utility and indicators}
    The utility dimension, denoted as \( \mathcal{U} \), evaluates how well synthetic data can support meaningful downstream analytical tasks compared to the real data. Therefore, the utility can be assessed across representative categories, including \textit{monitoring} (e.g., trend detection and system surveillance), \textit{predictive modeling}(e.g., training and validating machine learning models) and \textit{policy analysis} (e.g., \textit{`before-after'} evaluations, scenario simulations). In practice, utility should be assessed in a task-specific manner. For example, clustering utility can be evaluated by applying K-Means separately to real and synthetic data and comparing the resulting centroids using Euclidean distance. For predictive modeling, a common strategy is \textit{Train on Synthetic, Test on Real} (TSTR), which evaluates how well models trained on synthetic data generalize to real observations. The resulting performance is then compared to a baseline established by \textit{Train on Real, Test on Real} (TRTR), providing a reference for the predictive utility gap between synthetic and real data.

\section{Generative models for benchmarking}\label{sec:models}
  \subsection{Statistical models}
    Classical examples of generative models include Gaussian Mixture Models (GMMs), Gaussian copulas, and Bayesian networks (BN).
    GMMs assume that data is generated from a mixture of several Gaussian distributions and estimate parameters through expectation-maximization \citep{reynolds2015gaussian}. Real data is used to estimate the parameters of the distributions and then the synthetic data can be produced by sampling a component according to the learned mixture weights and drawing a sample from the corresponding Gaussian distribution.
    
    Gaussian Copulas separate the modeling of marginal distributions from their dependency structure. In the first step, each variable’s marginal distribution is estimated and transformed into a standard normal space via the probability integral transform. Then, a multivariate Gaussian distribution is fit to capture the dependencies between transformed variables \citep{nelsen2006introduction}. Synthetic data is generated by sampling from the fitted multivariate normal distribution and then inverting the transformation back to the original data scale using the estimated marginals.
    
    Bayesian Networks (BNs) represent the joint distribution of a set of variables using a directed acyclic graph (DAG), where each node corresponds to a variable and edges encode conditional dependencies. The structure of the graph is learned from data, and conditional probability tables (CPTs) are estimated for each node given its parents \citep{koller2009probabilistic}. Synthetic data is generated by ancestral sampling: variables are sampled in topological order based on the graph, using the learned CPTs.  
    
    % While these models are interpretable and effective in many applications, they typically rely on simplifying assumptions (e.g., Gaussianity, conditional independence) and may struggle to capture complex, high-dimensional, or non-linear patterns in real-world data such as mobility behaviors. Nevertheless, they provide strong and transparent baselines for evaluating the performance of deep generative models (DGMs). Moreover, the foundational concepts behind these classical models have inspired enhancements to modern deep learning approaches. For instance, GMM is incorporated into Conditional Tabular GANs (CTGAN) to guide training data sampling and mitigate mode collapse \citep{xu2019modeling}; copula-based transformations have been introduced to improve data normalization and enhance the transferability of DGMs \citep{jutras2024copula} and BN is leveraged to derive topological orderings that structure and constrain the autoregressive generation process in large language models (LLMs) for population synthesis tasks \citep{lim2025large}.

  % \subsection{Deep generative models}
    % Unlike classical models, DGMs leverage neural network architectures to flexibly approximate non-linear dependencies and latent structure in the data. In the context of synthetic data generation, DGMs offer the potential to produce realistic, diverse samples that preserve both global and local statistical properties of the original dataset.
    
  \subsection{Generative adversarial networks}
    Generative Adversarial Networks (GANs) are a class of deep generative models that learn to produce synthetic data by framing the generation process as a two-player minimax game between a generator and a discriminator \citep{goodfellow2014generative}. 
    Let \( G(z; \theta_G) \) be a generator that maps random noise \( z \sim p_z(z) \) to synthetic samples, and \( D(x; \theta_D) \) be a discriminator that outputs the probability that input \( x \) is a real trip rather than a synthetic one. The objective function is formulated as:
            \begin{equation}\label{eq:rq_o}    
                \min_{G} \max_{D} \mathbb{E}_{x \sim p_{\text{data}}(x)}[\log D(x)] + \mathbb{E}_{z \sim p_z(z)}[\log (1 - D(G(z)))]
            \end{equation}

    During training, the discriminator \( D \) tries to distinguish real trips \( x \sim p_{\text{data}} \) from generated ones \( G(z) \), while the generator \( G \) attempts to fool the discriminator by producing trips \( x' = G(z) \) that resemble real ones. 
    It is proven in \cite{goodfellow2014generative} that the dual problem in Eq.~\ref{eq:rq_o} is equivalent to minimizing the JSD between the generated distribution and the real distribution.
    However, the training suffers from issues such as vanishing gradients when minimizing \(\log (1 - D(G(z)))\), which can hinder effective learning. To address this, the Wasserstein GAN (WGAN) \citep{arjovsky2017wasserstein} was proposed replacing the Jensen–Shannon divergence with the Wasserstein-1 distance to provide smoother gradients and more stable training. This was further improved by adding a gradient penalty in WGAN-GP \citep{gulrajani2017improved}. In this study, WGAN-GP is adopted and evaluated due to its improved convergence properties and robustness. 
    
    \textbf{Conditional GANs (cGANs)} extend GAN framework by conditioning both \( G \) and \( D \) on additional side information \( c \):
            \begin{equation}
                \min_{G} \max_{D} \mathbb{E}_{x \sim p_{\text{data}}(x)}[\log D(x \mid c)] + \mathbb{E}_{z \sim p_z(z)}[\log (1 - D(G(z \mid c)))]
            \end{equation}

    In the transit context, \( c \) could represent features such as the intended origin station, travel time interval, or passenger category. Conditioning has the potential to improve the model’s ability to generate context-aware synthetic trips, increasing their behavioral plausibility and control. Conditional Tabular GAN (CTGAN) \citep{xu2019modeling}, which explicitly incorporates conditioning during training to better handle mixed data types and imbalanced distributions, falls into this category and is benchmarked in this study.
    
        % In the context of transit data, GANs aim to generate synthetic trips that are indistinguishable from real trips recorded in AFC data.

        % Although GANs are powerful, they present several challenges in mobility applications, including unstable training dynamics, difficulty in capturing sequential dependencies, and limited interpretability. Extensions such as recurrent GANs, trajectory GANs, and domain-constrained GANs have been proposed to address these issues, but require careful design to align with transit-specific constraints such as route feasibility and transfer logic.

  \subsection{Variational autoencoders}
    
    Variational Autoencoders (VAEs) are probabilistic generative models that learn a latent representation of data and generate new samples by decoding from this latent space \citep{kingma2013auto}. 

    A VAE consists of two neural networks: an encoder \( q_{\phi}(z \mid x) \), which approximates the posterior distribution over latent variables \( z \) given input data \( x \), and a decoder \( p_{\theta}(x \mid z) \), which reconstructs the input data from the latent space. The VAE is trained to maximize the evidence lower bound (ELBO) on the marginal log-likelihood of the data:
    \begin{equation}
        \log p_{\theta}(x) \geq \mathbb{E}_{q_{\phi}(z \mid x)}[\log p_{\theta}(x \mid z)] - \text{KL}(q_{\phi}(z \mid x) \| p(z))
    \end{equation}
    where \( \text{KL}(\cdot \| \cdot) \) denotes the KLD between the approximate posterior and the prior \( p(z) \), typically taken as a standard normal distribution \( \mathcal{N}(0, I) \).
    The first term encourages accurate reconstruction of the input data, while the second term regularizes the latent space to conform to the prior distribution, promoting smoothness and generalization.

    A \textbf{conditional VAE (cVAE)} extends VAE framework by conditioning both encoder and decoder on observed features \( c \), such as temporal context or user profiles:
    \begin{equation}
        \log p_{\theta}(x \mid c) \geq \mathbb{E}_{q_{\phi}(z \mid x, c)}[\log p_{\theta}(x \mid z, c)] - \text{KL}(q_{\phi}(z \mid x, c) \| p(z \mid c))
    \end{equation}

    For public transit applications, the input of VAEs and cVAEs may consist of structured features such as origin, destination, departure time, route characteristics, or user type. The decoder learns to generate synthetic trips that reflect these characteristics while capturing latent variability in the underlying behavior. Both VAE \citep{kingma2013auto} and cVAE \citep{xu2019modeling} are benchmarked in this study. 
    
  \subsection{Diffusion models}
    Diffusion models are a class of generative models that synthesize data by reversing a gradual noising process, learning to transform pure noise into structured samples through a sequence of denoising steps \citep{ho2020denoising}. These models have recently achieved state-of-the-art results in image and time-series generation and are now being explored in mobility domains for their ability to capture fine-grained temporal dynamics and distributional diversity.

    The diffusion process consists of two stages: a forward process (diffusion), where noise is incrementally added to the data, and a reverse process (generation), where the model learns to progressively denoise and recover the original data. Formally, the forward process defines a Markov chain \( q(x_t \mid x_{t-1}) \) that gradually adds Gaussian noise to a data sample \( x_0 \) over \( T \) time steps:
    \begin{equation}
    q(x_t \mid x_{t-1}) = \mathcal{N}(x_t; \sqrt{1 - \beta_t} \, x_{t-1}, \beta_t I)
    \end{equation}
    where \( \beta_t \in (0, 1) \) controls the noise schedule.

    The reverse process is parameterized by a neural network \( \epsilon_{\theta}(x_t, t) \), trained to predict and remove the added noise. The generative model approximates the reverse transition as:
    \begin{equation}
    p_{\theta}(x_{t-1} \mid x_t) = \mathcal{N}(x_{t-1}; \mu_{\theta}(x_t, t), \Sigma_{\theta}(x_t, t))
    \end{equation}

    Training involves minimizing a variational bound, which in practice reduces to a simplified denoising score matching objective:
        \begin{equation}
            \mathcal{L}_{\text{simple}}(\theta) = \mathbb{E}_{x_0, \epsilon, t} \left[ \left\| \epsilon - \epsilon_{\theta}(\sqrt{\bar{\alpha}_t} x_0 + \sqrt{1 - \bar{\alpha}_t} \, \epsilon, t) \right\|^2 \right]
        \end{equation}
    where \( \bar{\alpha}_t \) is the cumulative product of \( (1 - \beta_t) \), and \( \epsilon \sim \mathcal{N}(0, I) \).

    In the context of synthetic public transit trips, diffusion models can be applied to learn distributions over trip features or sequences, such as travel times, origins, or trajectory embeddings. Their iterative nature allows for modeling multimodal distributions and uncertainty in generation, which is valuable in replicating diverse mobility patterns.
    However, diffusion models present challenges in practice \citep{peng2024diffusion}. They they are computationally intensive due to the large number of sampling steps, and they require careful architectural design to incorporate domain-specific constraints. To this end, a multilayer perceptron (MLP)-based denoiser is customized and benchmarked in this study.  

  \subsection{Normalizing flow}
    Normalizing Flows (NFs) are a class of likelihood-based generative models that transform a simple base distribution into a complex target distribution through a series of \textit{invertible} and \textit{differentiable} mappings. 

    Let \( \mathbf{x} \in \mathbb{R}^d \) denote a real-valued data vector (e.g., a trip record), and \( \mathbf{z} \in \mathbb{R}^d \) be a latent variable sampled from a known prior \( p_{\mathbf{Z}}(\mathbf{z}) \), typically a standard multivariate normal distribution. A normalizing flow defines a sequence of \( K \) invertible functions \( f_k \) such that:
    \begin{equation}
        \mathbf{x} = f_K \circ f_{K-1} \circ \dots \circ f_1(\mathbf{z}) = f(\mathbf{z}), \quad \mathbf{z} = f^{-1}(\mathbf{x})
    \end{equation}

    The log-likelihood of a data point \( \mathbf{x} \) is given by the change-of-variables formula:
    \begin{equation}\label{equ:nf2}
        \begin{split}
            \log p_{\mathbf{X}}(\mathbf{x}) &= \log p_{\mathbf{Z}}(f^{-1}(\mathbf{x})) 
                + \sum_{k=1}^K \log \left| \det \left( \frac{\partial f_k^{-1}}{\partial \mathbf{h}_k} \right) \right|,\\
            & \text{where } \mathbf{h}_k = f_k^{-1} \circ \dots \circ f_1^{-1}(\mathbf{x})
        \end{split}
    \end{equation}
    the Jacobian determinant quantifies the volume change introduced by each transformation.

    In our implementation, we adopt a RealNVP-style flow~\citep{dinh2016density} using affine coupling layers. Each coupling layer splits the input \( \mathbf{x} = [\mathbf{x}_A, \mathbf{x}_B] \) and applies the following transformation:

    \[
        \begin{aligned}
            \mathbf{y}_A &= \mathbf{x}_A, \\
            \mathbf{y}_B &= \mathbf{x}_B \odot \exp(s(\mathbf{x}_A)) + t(\mathbf{x}_A),
        \end{aligned}
    \]    
    where \( s(\cdot) \) and \( t(\cdot) \) are neural networks producing scaling and translation parameters, and \( \odot \) denotes element-wise multiplication. The triangular Jacobian structure enables efficient training via exact likelihood computation.
    
    To synthesize trips, the NF model is trained by maximizing the log-likelihood on real trip vectors, consisting of normalized continuous features and embedded or one-hot categorical features. Synthetic data is generated by sampling \( \mathbf{z} \sim \mathcal{N}(0, I) \) and applying the forward transformation \( \mathbf{x} = f(\mathbf{z}) \). Outputs are postprocessed to decode categorical variables (e.g., origin, destination) from softmax outputs or discrete embeddings.
    
    % While NFs offer exact density modeling and invertibility, they may face challenges with discrete or high-cardinality categorical features, common in mobility datasets. Addressing these issues often requires embedding layers, dequantization strategies, or hybrid architectures.

  \subsection{LLM as generators}
    Large Language Models (LLMs), such as GPT and BERT variants, have demonstrated impressive generative capabilities across domains including text, code, and structured data \citep{raiaan2024review}. More recently, LLMs have been explored for synthetic data generation in tabular and sequential settings, including human mobility modeling, by treating structured records as tokenized sequences and learning the joint distribution through language modeling objectives \citep{borisov2022language,lim2025large}.

    In the context of public transit, synthetic trips can be formulated as sequences of discrete tokens, where each token represents a feature or event, such as origin station, departure time, travel mode, or transfer point. A synthetic trip \( T \) can be expressed as a sequence:
    \begin{equation}
        T = \{ w_1, w_2, \dots, w_n \}
    \end{equation}
    where each token \( w_i \in \mathcal{V} \) is drawn from a vocabulary \( \mathcal{V} \) of possible trip elements (e.g., station codes, time bins, user types).

    LLMs are trained using an autoregressive objective that models the probability of the next token given the preceding ones:
    \begin{equation}
        P(T) = \prod_{i=1}^{n} P(w_i \mid w_1, \dots, w_{i-1})
    \end{equation}

    Fine-tuning or prompt engineering can incorporate contextual conditions such as user type, day of the week, or policy scenario. In such cases, the model can be conditioned on a prefix \( c \), modifying the generation as:
    \begin{equation}
        P(T \mid c) = \prod_{i=1}^{n} P(w_i \mid c, w_1, \dots, w_{i-1})
    \end{equation}

    This generative approach enables flexible sampling and customization, as well as natural integration of heterogeneous data. For example, trip chains, passenger attributes, and even travel policies can be embedded within token sequences or prompt templates. In this study, we customize and benchmark the generative LLM proposed by \cite{zhao2025tabula}. 

    % However, using LLMs for synthetic transit trip generation presents several limitations. First, ensuring that generated trips comply with system rules (e.g., feasible transfers, spatial alignment) is non-trivial. Second, discretization of continuous features (e.g., time, fare) may lead to information loss or binning artifacts. Furthermore, LLMs may overfit dominant patterns or generate implausible behaviors not grounded in system logic.

    % Despite these challenges, LLM-based approaches offer new opportunities for generating semantically rich, customizable synthetic data with minimal feature engineering. Ongoing research aims to enhance controllability and incorporate domain-specific constraints directly into the generative process.
        
  \subsection{Privacy-enhanced models}
    Differential Privacy (DP) \citep{dwork2006calibrating} provides a formal mathematical framework for quantifying the privacy guarantees of algorithms that release information about datasets. A randomized algorithm \( \mathcal{A} \) satisfies \(\varepsilon\)-differential privacy if, for any two neighboring datasets \( D \) and \( D' \) differing in at most one record, and for any measurable subset of outputs \( S \subseteq \text{Range}(\mathcal{A}) \), the following condition holds:
    \begin{equation}\label{eq:epsilon-dp}
        \Pr[\mathcal{A}(D) \in S] \leq e^{\varepsilon} \cdot \Pr[\mathcal{A}(D') \in S]
    \end{equation}
    Here, \( \varepsilon > 0 \) is the \textit{privacy budget}, with smaller values implying stronger privacy protection. This ensures that the inclusion or exclusion of a single individual's data has a limited influence on the algorithm's output.

    A more relaxed version, known as \((\varepsilon, \delta)\)-differential privacy, allows for a small probability \( \delta \) that the privacy guarantee is violated:
    \begin{equation}\label{eq:eps-delta-dp}
        \Pr[\mathcal{A}(D) \in S] \leq e^{\varepsilon} \cdot \Pr[\mathcal{A}(D') \in S] + \delta
    \end{equation}
    This relaxation is often necessary in deep learning-based mechanisms \citep{abadi2016deep} such as the Gaussian mechanism or teacher-student aggregation schemes. In our study, we consider both definitions depending on the generative model used.

    To benchmark privacy-preserving enhanced synthetic data generation, we include two representative models that incorporate differential privacy guarantees using distinct methodological approaches: statistical Priv-BN \citep{zhang2017privbayes} and neural PATE-GAN \citep{jordon2018pate}.
    \textbf{Priv-BN} (PrivBayes) constructs a differentially private Bayesian Network to approximate the joint distribution of tabular data. It iteratively learns a set of low-dimensional conditional distributions by building a directed acyclic graph over the attributes. To achieve \(\varepsilon\)-differential privacy, Laplace noise is added to the computation of marginal and conditional probabilities during both the structure learning and parameter estimation stages. The final synthetic dataset is generated by sampling from the noisy Bayesian network. This approach offers interpretability and strong privacy guarantees, though it may struggle with capturing complex, high-dimensional dependencies.
    \textbf{PATE-GAN} integrates the Private Aggregation of Teacher Ensembles (PATE) framework into a GAN architecture to achieve \((\varepsilon, \delta)\)-differential privacy. The dataset is split into disjoint subsets, each used to train a teacher classifier (discriminator). Noisy majority voting across these teachers provides labels used to train a student model, ensuring differential privacy via the Gaussian mechanism. This student model is then used to guide the discriminator in adversarial training. 
    The generator learns to produce data that can fool the discriminator while remaining differentially private through the PATE aggregation. PATE-GAN is particularly suited to high-dimensional data but introduces trade-offs between utility and the amount of noise required to protect privacy.
 
\section{Case study and results}\label{sec:case}
  \subsection{Data description}
    The Mass Transit Railway (MTR) in Hong Kong accommodates an average of 5 million passenger trips per day, with the vast majority paid through the Octopus Card, a contactless smart card embedded within the automatic fare collection (AFC) system. Key trip details such as anonymized card IDs, card type, tap-in and tap-out times/stations, and fare deductions are recorded in the transaction data. The ubiquity, comprehensive coverage, and high spatiotemporal resolution of such data have made it a valuable resource for advancing transit modeling \citep{nassir2019strategy, mo2021calibrating}, operational intelligence \citep{halvorsen2020demand}, passenger behavior analysis \citep{ma2020behavioral}, and public policy evaluation \citep{wu2025data}. However, access to this data is restricted due to privacy concerns, limiting its potential for open research and collaboration.

\begin{figure}[h]
        \centering
        \includegraphics[width=0.8\linewidth]{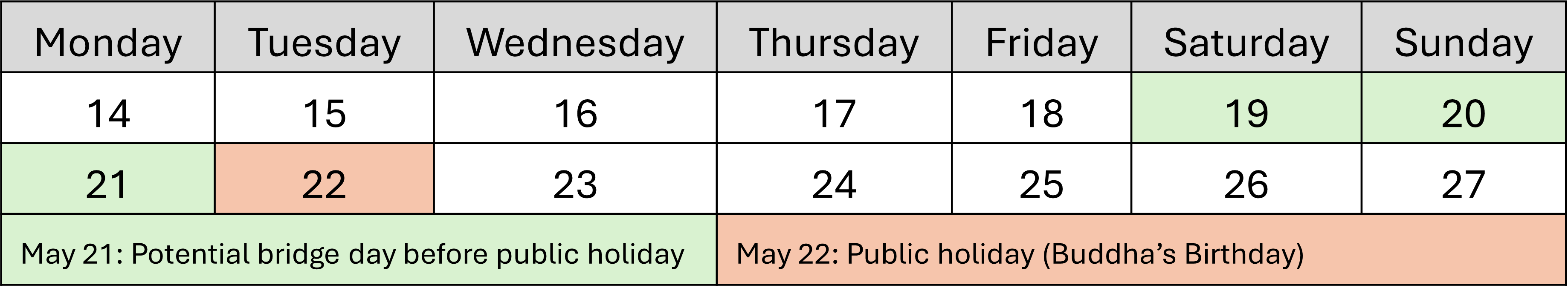}
        \caption{Calendar breakdown of selected data period (May 14–27, 2018).}
        \label{fig:date}
    \end{figure}

    In this study, we use Octopus card data as the real reference dataset and evaluate the performance of various generative models in producing synthetic trip data. We aim to assess whether synthetic data can adequately represent real data, support similar downstream applications, and offer reduced privacy risks for data sharing. 
    % Specifically, we benchmark different generative models to determine their overall performance, examine their strengths and limitations in replicating transit trip patterns, and identify persistent challenges and open questions in the generation of realistic and privacy-preserving synthetic transit data. 
    The real dataset used in this study consists of anonymized Octopus card transactions collected over a two-week period from May 14 to May 27, 2018 (Figure~\ref{fig:date}). This period was deliberately selected to capture a diverse range of travel patterns influenced by regular weekdays, weekends, and a public holiday (May 22, Buddha's birthday). 
    The selected period allows us to observe several anticipated variations in travel behavior, including typical weekday commuting patterns, regular weekend travel, holiday-related behaviors, and differences between normal and holiday-affected workdays or weekends. 
    % This temporal diversity is critical for assessing the ability of generative models to capture context-sensitive mobility dynamics in synthetic data.

\begin{figure}[h]
        \centering
        \includegraphics[width=0.8\linewidth]{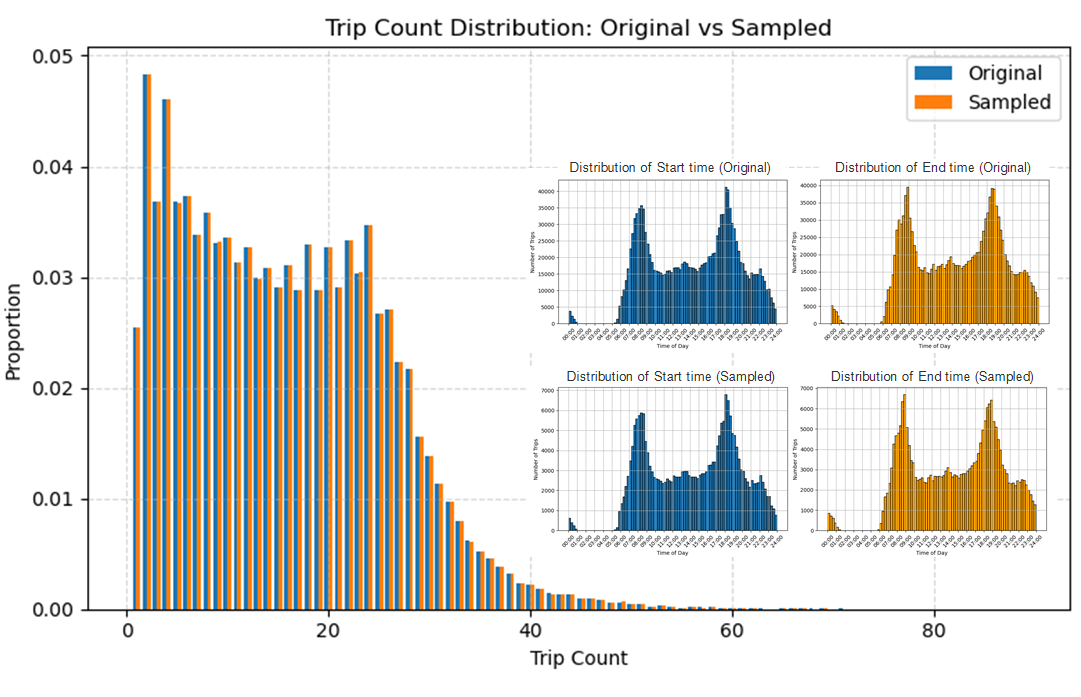}
        \caption{The sampled datasets preserve the original distribution of trip counts, start time and end time.}
        \label{fig:datadistribution}
    \end{figure}

    From the raw dataset, we extracted three disjoint subsets for model training, testing, and validation. Each record in the extracted data includes the following features: anonymized passenger ID, trip origin, destination, start time, end time, and the corresponding day of the week as conditional information. The training set consists of 9,000 passengers with a total of 140,725 trips, the testing set includes 3,011 passengers with 47,283 trips, and the validation set comprises 2,985 passengers with 46,395 trips. The sampling procedure was designed to preserve the original distributions of trip count per passenger, as well as the temporal characteristics of start time and end time. The distributional consistency is illustrated in Figure~\ref{fig:datadistribution}. 
    
  \subsection{Calculation and normalization of evaluation scores}
    To ensure fairness in comparison, all models are trained on the same input data and are required to generate an equal number of synthetic trips, denoted as (\(T_{syn}\)). No post-processing is applied to the generated data, allowing for a direct assessment of raw model performance.

    We first assess record-level representativeness by applying logical consistency checks to each generated trip. Specifically, we examine whether the generated origin-destination (OD) pairs exist in the original data, whether the start time precedes the end time, and whether time values fall within the valid range (0–1440 minutes, as all times are encoded in minutes past midnight). Records that fail any of these checks are counted as invalid and denoted by \(T_{syn_f}\). The record-level representativeness score, \(\mathcal{R}_r\), is then defined as the proportion of logically consistent records:
    \[
        \mathcal{R}_r = 1-\frac{T_{syn_f}}{T_{syn}}
    \]
    This metric reflects the model’s ability to generate plausible individual records consistent with real-world constraints.

    Population-level representativeness is evaluated using three statistical distance measures: Kullback–Leibler divergence (KLD), Jensen–Shannon divergence (JSD), and Earth Mover’s Distance (EMD) as introduced in subsection~\ref{subsec:rindicator}, denoted as \( \mathcal{R}_{p_k} \), \( \mathcal{R}_{p_j} \), and \( \mathcal{R}_{p_e} \), respectively. 
    Similarly, group-level representativeness is assessed by comparing distributions conditioned on the day of the week, using the same three metrics: \( \mathcal{R}_{g_k} \), \( \mathcal{R}_{g_j} \), and \( \mathcal{R}_{g_e} \).
    To ensure comparability across models, we apply \textit{min-max normalization} to each metric across all models. The final group- and population-level representativeness scores are then computed as the average of the normalized metrics:
        \[
            \mathcal{R}_g = \text{Average}(\mathcal{R}_{g_k}, \mathcal{R}_{g_j}, \mathcal{R}_{g_e}), \quad
            \mathcal{R}_p = \text{Average}(\mathcal{R}_{p_k}, \mathcal{R}_{p_j}, \mathcal{R}_{p_e})
        \]

    The overall representativeness score for a given model \( m \) is computed as the average of its record-level, group-level, and population-level scores:
        \[
            \mathcal{R}^m = \text{Average}(\mathcal{R}_r^m, \mathcal{R}_g^m, \mathcal{R}_p^m)
        \]

    Privacy evaluation is performed using membership inference attacks (MIA) and k-nearest neighbor (k-NN) analysis, as detailed in Algorithm \ref{alg:mia_rf} and \ref{alg:knn_privacy}. 
    Higher membership probability indicates greater privacy risk; thus, the mean MIA probability of each model is normalized to \(p_{mia}\), and the record-level privacy score is defined as \(\mathcal{P}_r = (1-p_{mia})\). 
    In the k-NN analysis, a larger distance ratio corresponds to lower privacy risk. The resulting distance ratios are normalized to obtain the population-level \(\mathcal{P}_p\) and group-level \(\mathcal{P}_g\) privacy scores. The overall privacy score for model \( m \) is computed as the average of its record-level, group-level, and population-level scores:
        \[
            \mathcal{P}^m = \text{Average}(\mathcal{P}_r^m, \mathcal{P}_g^m, \mathcal{P}_p^m)
        \]

    The utility evaluation is task-specific. In this study, we assess two downstream tasks: clustering and prediction.
    To assess clustering utility, K-Means clustering is applied separately to real and synthetic datasets, and the average minimum distance between their centroids is computed. A smaller distance indicates higher clustering utility.
    For predictive utility, a \textit{Train on Synthetic, Test on Real} (TSTR) evaluation is conducted by training a Gradient Boosting Regressor with \(k\)-fold cross-validation on both datasets and evaluating on real test folds. The difference in predictive accuracy, measured by Mean Absolute Error (MAE) and Root Mean Squared Error (RMSE), between TSTR and a \textit{Train on Real, Test on Real} (TRTR) baseline quantifies predictive utility. Smaller differences indicate better preservation of modeling utility.
    All utility indicators are first normalized across models: \(d_c\) (clustering distance), \(d_{mae}\), and \(d_{rmse}\) (prediction differences). The scores are defined as: \(\mathcal{U}_c^m = 1-d_c\) (clustering score),  \(\mathcal{U}_p^m = \text{Average}(1-d_{mae}, 1-d_{rmse})\) (prediction score), the overall utility score is computed as:
    \[
        \mathcal{U}^m = \text{Average}(\mathcal{U}_c^m , \mathcal{U}_p^m).
    \]
    
    It is important to note that all scores shown in the following figures are \textbf{min-max normalized} across models; therefore, lower values indicate comparatively weaker performance in a given dimension relative to other models, rather than poor absolute performance. 
    
  \subsection{Benchmarking results}  
    We present the benchmarking results of the proposed evaluation framework applied to a wide range of generative models as introduced in section~\ref{sec:models}. To begin, we compare all benchmarked methods across the three evaluation dimensions. 
    % This broad comparison provides an overall view of model performance before turning to a more detailed analysis of selected representative models.
    % In this study, we evaluate a diverse set of generative models for synthetic trip data generation as introduced in section~\ref{sec:models}. These include statistical models: Gaussian Mixture Model (GMM), Gaussian Copula (GC), and Bayesian Network (BN); deep generative models: WGAN-GP, CTGAN, VAE, cVAE, diffusion model, normalizing flow (NF), and a Large Language Model (LLM)-based generator; privacy-enhanced models: PrivBayes Network (Priv-BN) and PATE-GAN, both designed to provide differential privacy guarantees.
   \subsubsection{Evaluation across all methods}
\begin{figure}[!h]
        \centering
        \includegraphics[width=1\linewidth]{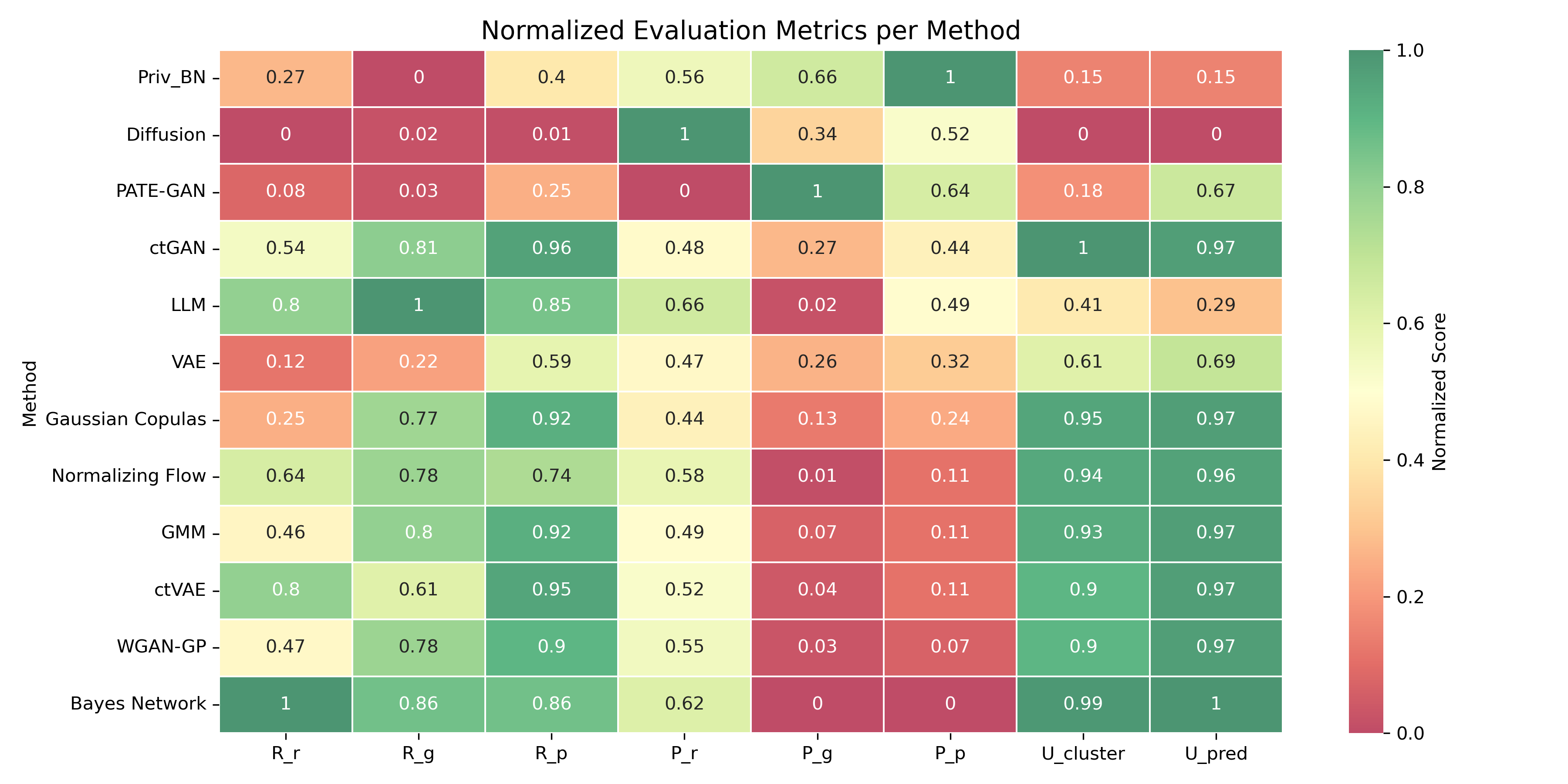}
        \caption{Normalized evaluation metrics per method including representativeness (\(\mathcal{R}_r,\mathcal{R}_g, \mathcal{R}_p\)), privacy (\(\mathcal{P}_r,\mathcal{P}_g, \mathcal{P}_p\)), and utility (\(\mathcal{U}_{cluster},\mathcal{R}_{pred}\)).}
        \label{fig:heatmap}
    \end{figure}

    Figure~\ref{fig:heatmap} presents the normalized evaluation metrics across all methods, revealing that no single model achieves uniformly strong performance across representativeness, privacy, and utility. The substantial variation across record-, group-, and population-level metrics highlights that relying on a single evaluation level may obscure important performance differences, reinforcing the importance of multi-level assessment. 

\begin{figure}[h!]
        \centering
        \includegraphics[width=1\linewidth]{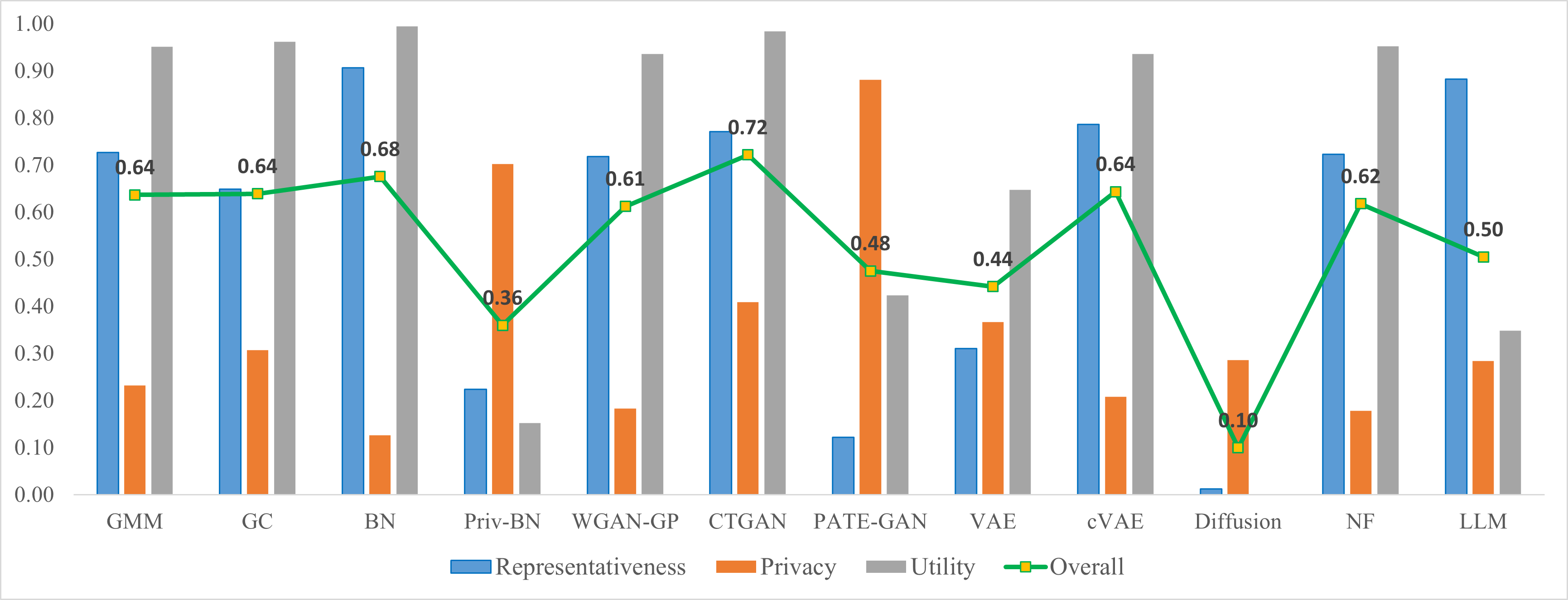}
        \caption{Overall performance per dimension of benchmarking models.}
        \label{fig:overall}
    \end{figure}

    Figure~\ref{fig:overall} presents the overall score for each dimension, along with the average across representativeness, privacy, and utility for each method. 
    Deep generative models (DGMs) do not surpass conventional statistical methods in overall performance, although they exhibit a slight advantage in privacy protection. A more promising direction is to improve DGM performance by leveraging strengths from conventional approaches. Incorporating GMM-based sampling to guide training and mitigate mode collapse, CTGAN is a good example to show the potential of this strategy. As a result, it achieves the highest overall performance, while ranking just behind Priv-BN and PATE-GAN in privacy protection. But these two privacy-enhanced models failed in the evaluation of representativeness and utility.

    Conditioned VAE (cVAE) demonstrates better overall performance than the unconditioned counterparts (VAE) as expected. However, the inclusion of additional side information in models like cVAE can lead to increased privacy leakage risks relative to VAE. Therefore, selecting an appropriate model requires careful consideration of the trade-off between performance and privacy in practical applications. Notably, the Normalizing Flow model achieved performance on par with cVAE, highlighting its strong potential as a foundation for further development in synthetic trip generation.  

    Interestingly, the LLM attains high representativeness but relatively low utility, diverging from the general trend observed in other models where utility scores are often aligned with representativeness. This suggests that preserving distributional similarity does not necessarily guarantee strong downstream analytical performance. The results of LLM generator reinforce the importance of evaluating synthetic data beyond representativeness, as relying solely on distributional similarity metrics risks overestimating practical utility. These findings underscore again the value of a comprehensive, multi-dimensional evaluation framework.
    
    It should be pointed out that to ensure comparability, all models in this study were trained on identically preprocessed data, in which categorical variables were one-hot encoded and continuous variables were min–max normalized. While this setting is consistent across models, it is not equally well-suited to every generative architecture. That is why diffusion model struggled. Diffusion models inject Gaussian noise directly into the feature space and iteratively denoise, which can disrupt the sparsity and mutual exclusivity inherent in one-hot representations, making it challenging to recover valid categorical structures. 
    
  \subsubsection{Detailed comparative analysis} 
    To provide a more concrete illustration of the evaluation framework, we conduct a detailed comparative analysis of four representative models: Bayes Network (BN), Priv-BN, CTGAN, and PATE-GAN. These models are selected as they span the key categories considered in this study: a statistical model (BN), a deep generative model (CTGAN), and their respective privacy-enhanced variants (Priv-BN and PATE-GAN). By comparing these models in greater detail, we highlight not only their relative performance across representativeness, privacy, and utility, but also the trade-offs that arise between fidelity and privacy when privacy-preserving mechanisms are incorporated.
    
\begin{figure}[h!]
        \centering
        \includegraphics[width=0.5\linewidth]{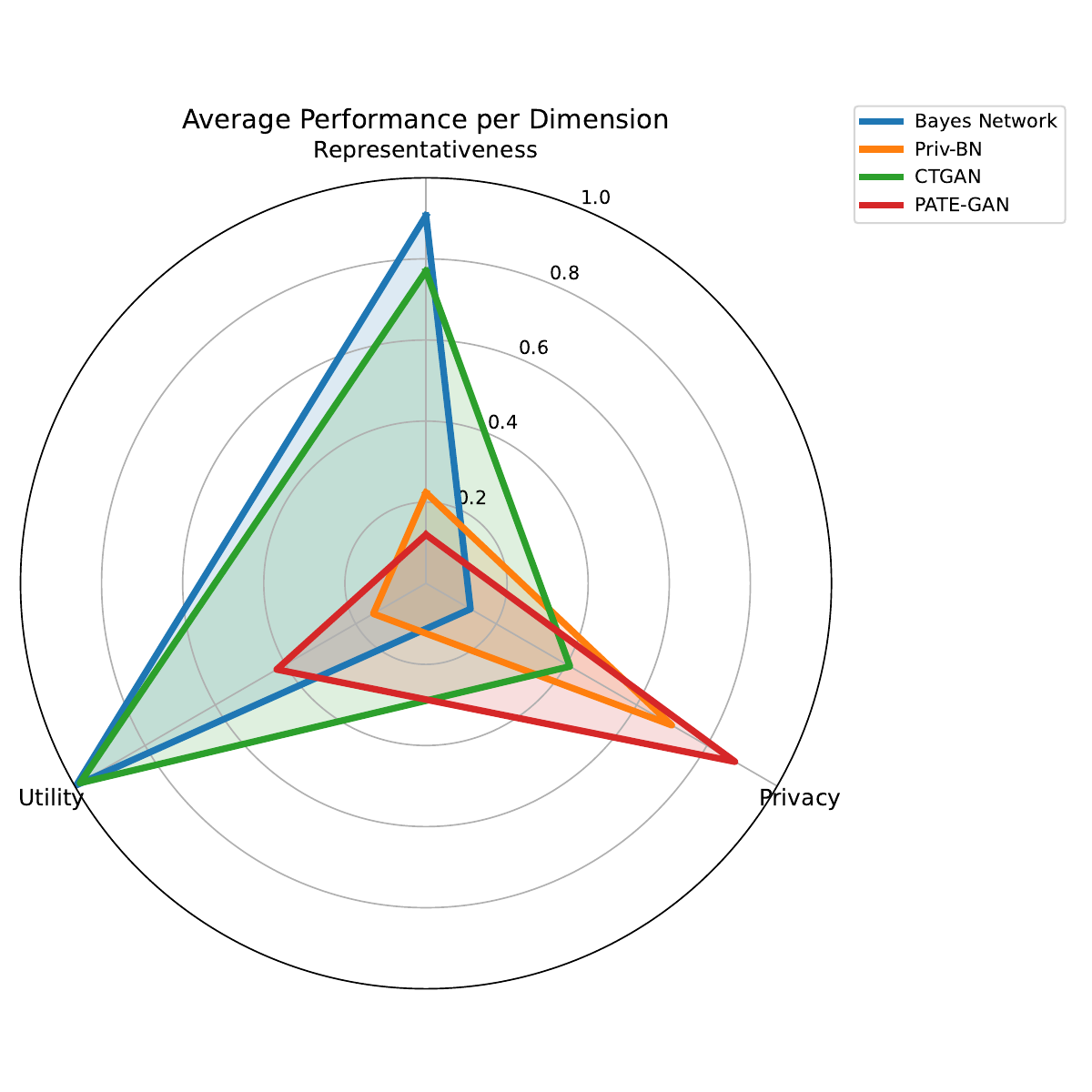}
        \caption{Average performance per dimension of Bayes Network, Priv-BN, CTGAN, and PATE-GAN.}
        \label{fig:radar}
    \end{figure}

    Figure~\ref{fig:radar} shows the average performance per dimension for the four selected models, clearly illustrating the fundamental trade-off between utility and privacy: gains in one dimension often come at the expense of the other. Models with high utility, such as BN and CTGAN, tend to exhibit weaker privacy performance, indicating a greater risk of information leakage when the synthetic data closely mirrors the real data distribution. In contrast, privacy-enhanced models like Priv-BN and PATE-GAN achieve stronger privacy protection but generally at the cost of representativeness and downstream analytical utility.

\begin{figure}[!h]
        \centering
        \includegraphics[width=0.9\linewidth]{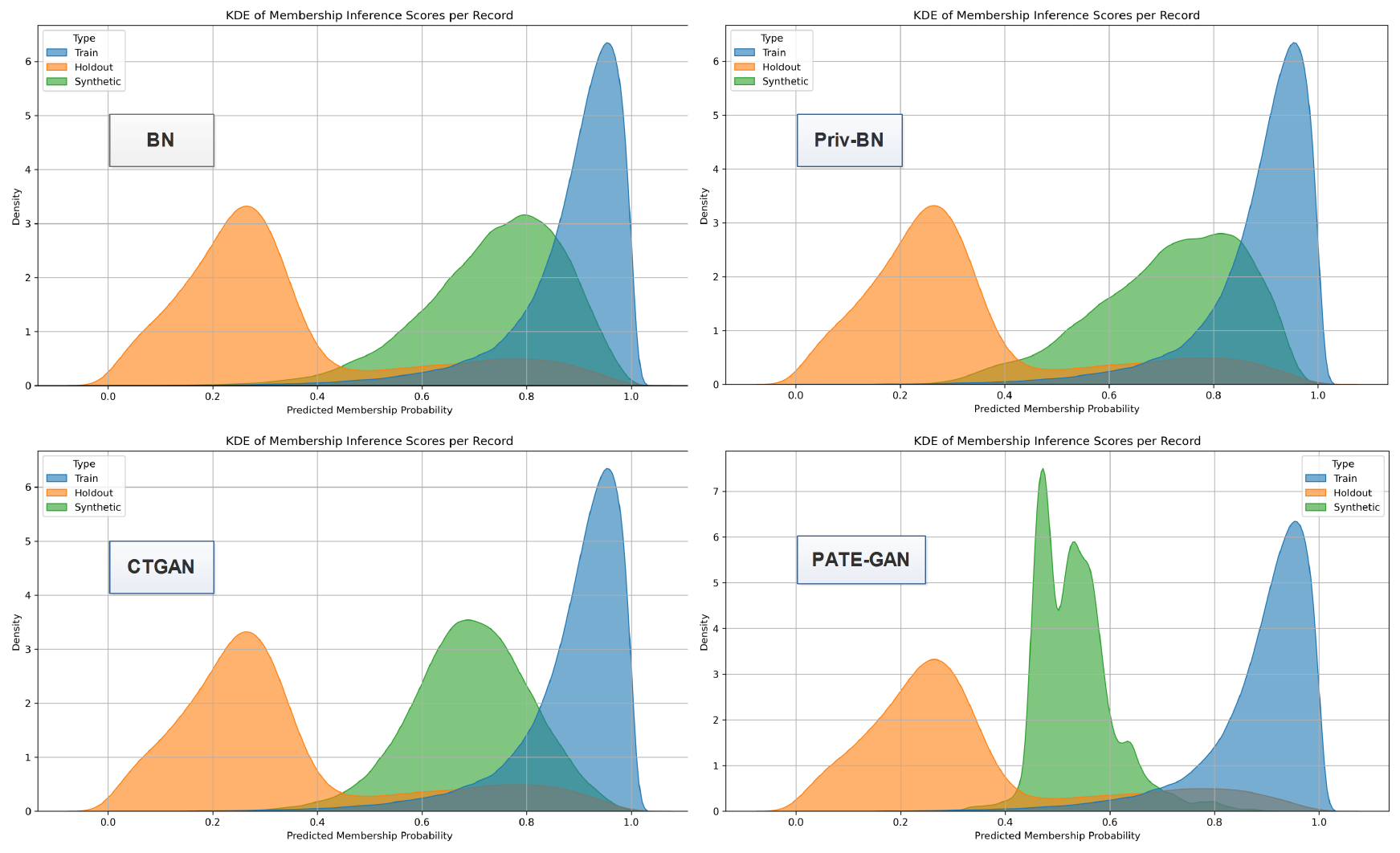}
        \caption{Privacy comparison of Bayes Network, Priv-BN, CTGAN, and PATE-GAN. The kernel density estimation (KDE) of membership inference attack (MIA) scores for synthetic records generated by each model are compared against the scores of real training and holdout (testing) records.}
        \label{fig:privacy_example}
    \end{figure}

    Figure~\ref{fig:privacy_example} presents the distribution of membership inference attack (MIA) scores for real training, holdout, and synthetic records across the four models. The kernel density estimation (KDE) curves illustrate how closely synthetic records resemble training or holdout data, which serves as an indicator of potential privacy leakage. The KDE distributions of synthetic records (green) overlap more strongly with those of the real training records (blue) than with holdout records (orange) for model BN and CTGAN. This suggests a higher risk of memorization and privacy leakage, as synthetic samples may reveal information about the training set. Compared to BN, the synthetic distribution of Priv-BN shifts slightly away from the training data, reducing overlap. This indicates that the differential privacy mechanism provides some mitigation of leakage, although at the expense of sample utility. The synthetic KDE of PATE-GAN is more distinct and separated from the training distribution, showing lower overlap with training data. This demonstrates stronger protection against membership inference, consistent with the model’s design to incorporate privacy-preserving aggregation.

\begin{figure}[!h]
        \centering        
        \includegraphics[width=\linewidth]{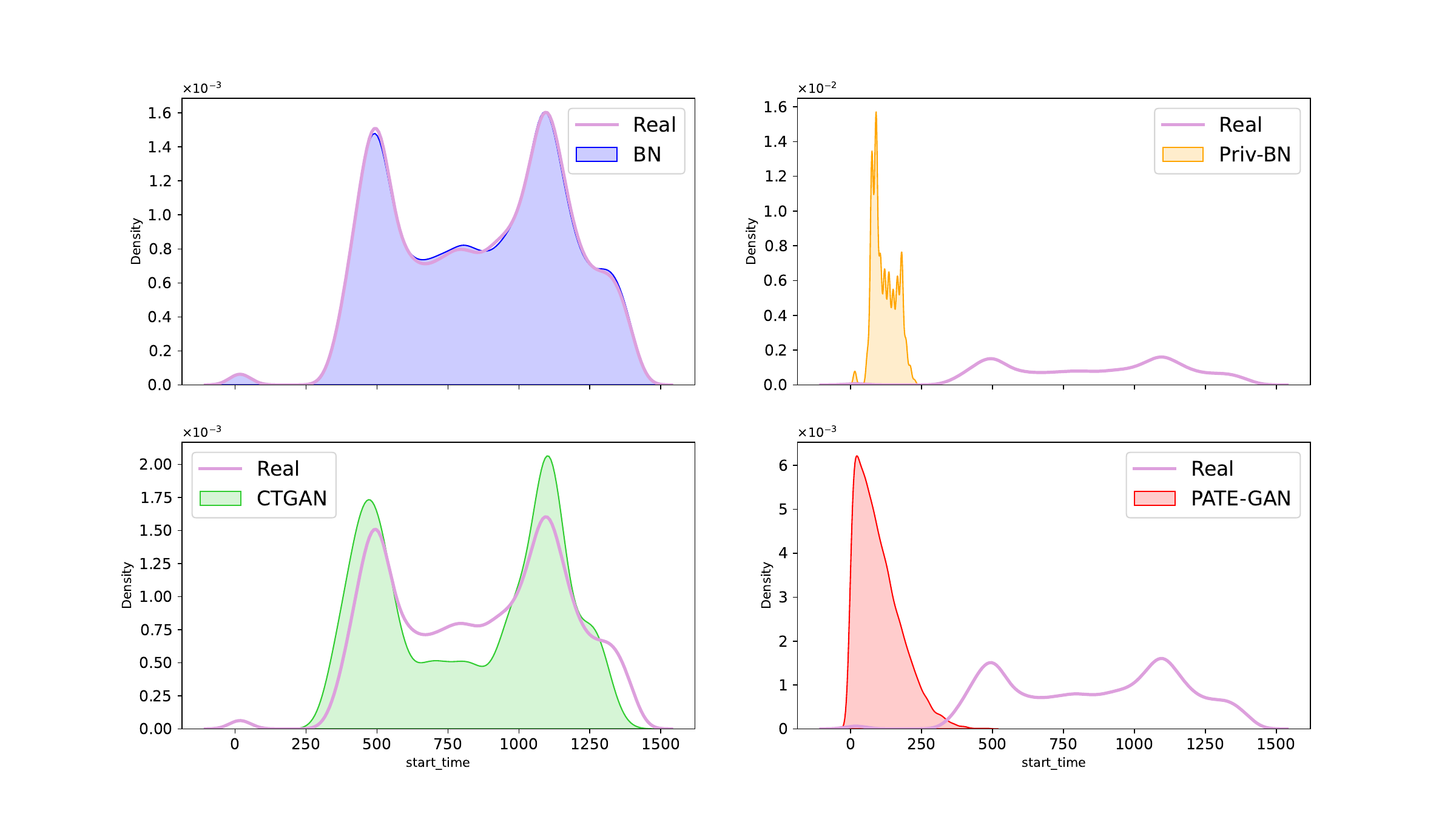}
        \caption{Representativeness comparison of Bayes Network, Priv-BN, CTGAN, and PATE-GAN. The kernel density estimates (KDE) of trip start times from the synthetic data generated by these models are compared against those from the real test data.}
        \label{fig:repre_example}
    \end{figure}

    Figure~\ref{fig:repre_example} shows the kernel density estimation (KDE) of trip start times in the real dataset compared with synthetic data generated by the four models. BN and CTGAN capture the main distributional patterns of real trip start times well, with KDE curves closely aligned to the real data. This indicates strong representativeness at the distributional level. Priv-BN shows a clear deviation from the real distribution, with synthetic trips concentrated at unrealistic early start times, reflecting the utility loss introduced by privacy constraints. PATE-GAN similarly fails to capture the multi-modal structure of real start times, generating distributions concentrated in unrealistic ranges.

\begin{figure}[!h]
        \centering
        \includegraphics[width=0.8\linewidth]{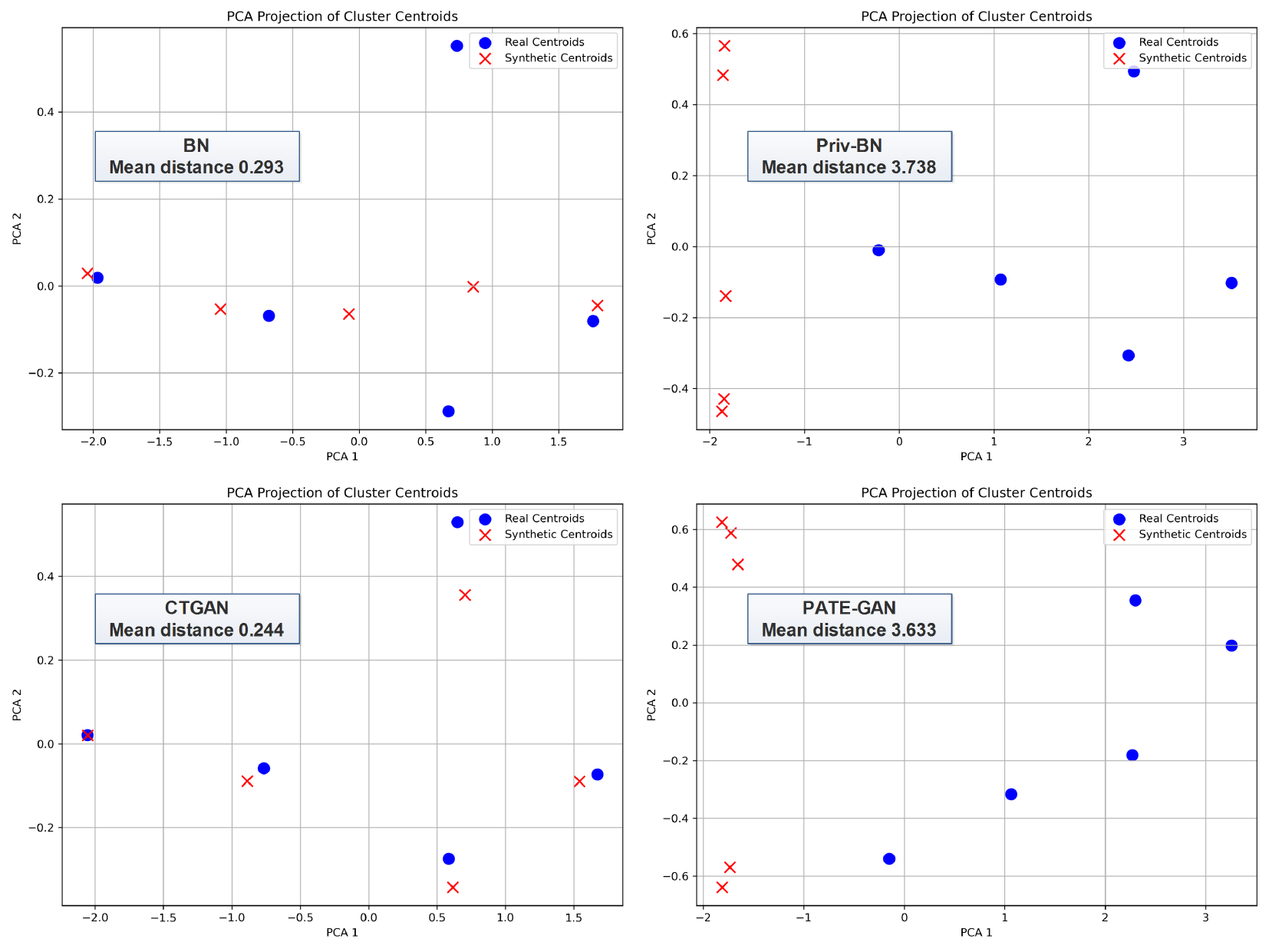}
        \caption{Utility comparison in the clustering task of Bayes Network, Priv-BN, CTGAN, and PATE-GAN. The evaluation is based on the average minimum distance between real and synthetic cluster centroids. Cluster centroids from the real and synthetic datasets are visualized in two dimensions using Principal Component Analysis (PCA).}
        \label{fig:cluster_example}
    \end{figure}

    Figure~\ref{fig:cluster_example} visualizes the cluster centroids from the real and synthetic datasets generated by the four models, projected into two dimensions using Principal Component Analysis (PCA). Consistent with the representativeness results, BN and CTGAN produce centroids that are closer to those of the real data, suggesting that their synthetic datasets are more suitable for clustering analysis. In contrast, the centroids from Priv-BN and PATE-GAN deviate substantially, indicating limited utility for clustering tasks.

    The detailed comparison of BN, Priv-BN, CTGAN, and PATE-GAN reflects the broader patterns observed across all benchmarked methods. Taken together, the results underscore the importance of a multi-dimensional, multi-level evaluation when assessing generative models. While developed in the context of synthetic trips, the proposed framework is equally relevant to other applications such as trajectory generation, where outputs must also be assessed across multiple dimensions and levels. To support transparency and reproducibility, we make the complete preprocessing and training code available for reference and examination by other researchers.
    
    % Models without explicit privacy mechanisms, such as BN and CTGAN, achieve stronger representativeness and utility but exhibit higher risks of privacy leakage. In contrast, their privacy-enhanced counterparts, Priv-BN and PATE-GAN, provide stronger protection against inference attacks but at the cost of reduced fidelity to real data. These findings underscore the inherent trade-offs between privacy and utility in synthetic data generation, which are further examined in the comprehensive evaluation across all models.
    
\section{Conclusion}\label{sec:conclusion}   
    This study introduced a comprehensive, multi-dimensional framework for evaluating synthetic trip data, spanning representativeness and privacy across record-, group-, and population-levels, as well as utility of different downstream tasks. Applying this framework to a diverse set of statistical and deep generative models revealed that no single approach dominates across all dimensions. Statistical models, such as Bayesian Networks, deliver strong representativeness and utility but carry substantial privacy risks, while privacy-enhanced variants like Priv-BN offer improved protection at the cost of data fidelity and analytical value. Among deep generative models, CTGAN achieved the most balanced performance, demonstrating that integrating conventional modeling components, such as GMM-based sampling, can enhance both fidelity and privacy. In contrast, formal differential privacy mechanisms, exemplified by PATE-GAN, provided strong privacy guarantees but significantly reduced representativeness and utility.

    Overall, our results highlight the inherent trade-offs between privacy and utility in synthetic data generation for public transit applications, and demonstrate the necessity of multi-dimensional, multi-level evaluation to avoid misleading conclusions from single-metric or single-level assessments. The proposed framework provides a transparent and reproducible basis for benchmarking generative models, guiding the selection and adaptation of methods for responsible deployment in transport research and policy analysis.

    While the evaluation framework is broadly applicable, several limitations should be acknowledged. First, the results reflect a single real-world dataset with fixed preprocessing, which may influence model performance rankings. Second, the scope of evaluation, although multi-dimensional, does not exhaust all possible measures of privacy risk or domain-specific utility. Finally, hyperparameter tuning was conducted within practical constraints, and alternative settings might yield different trade-offs. Building on our current benchmarking findings, we plan to investigate hybrid architectures that integrate the strengths of statistical and deep generative approaches. We also will extend the framework to other transport datasets that contain more sensitive personal information. 
    
\section*{Data and Code Availability}
    The code supporting the findings of this study is available at 
\href{https://github.com/DarinyWu/Syn-Data-Eva-Framework}{GitHub}. Due to the sensitive nature of the data used in this study, the data will remain confidential and cannot be publicly shared. However, access may be granted upon reasonable request by contacting authors. 

\section*{Acknowledgements}
    We would like to acknowledge TRENoP research centers in Sweden for their funding supports. This work was also in part financially supported by Digital Futures.
    During the preparation of this work the authors used ChatGPT 4o in order to check the English grammar. After using this tool, the authors reviewed and edited the content as needed and take full responsibility for the content of the publication.
% Thanks to .
% ..

% %% The Appendices part is started with the command \appendix;
% %% appendix sections are then done as normal sections
% \appendix

% \section{Appendix title 1}
% %% \label{}

% \section{Appendix title 2}


\bibliographystyle{elsarticle-harv}
\bibliography{synthetictrips}



%% For example:

%% \bibitem[Aladro et al.(2015)]{Aladro15} Aladro, R., Martín, S., Riquelme, D., et al. 2015, \aas, 579, A101

%%\end{thebibliography}

\end{document}